\documentclass[runningheads]{llncs}

\usepackage{eccv}

\usepackage{eccvabbrv}

\usepackage{tcolorbox}
\usepackage{graphicx}
\usepackage{booktabs}
\usepackage[numbers]{natbib}
\usepackage{array}
\usepackage{multirow}
\usepackage{subcaption}

\definecolor{CommentGreen}{rgb}{0.121,0.636,0.121} 
\definecolor{CommentBlue}{rgb}{0,0,1}
\definecolor{BackToBlack}{rgb}{0,0,0}
\definecolor{CommentPink}{rgb}{1,0.2,0.5}
\usepackage{amsmath}
\usepackage{amsfonts}

\DeclareMathOperator*{\argmin}{arg\,min}
\DeclareRobustCommand{\newstuff}[1]{\textcolor{BackToBlack}{#1}}
\usepackage{float}

\usepackage[accsupp]{axessibility}  

\usepackage{hyperref}

\usepackage{orcidlink}

\begin{document}
\crefname{section}{Appendix}{Appendices}

\title{CoReLIN: Constraint-based Reasoning for Zero-shot Lifelong Interactive Navigation} 

\titlerunning{CoReLIN}

\author{Apoorva Vashisth\inst{1}$^\ast$\orcidlink{0009-0005-3741-9053} \and
Manav Kulshrestha\inst{1}\orcidlink{0009-0004-3266-8328} \and
Pranav Bakshi\inst{2}\orcidlink{0009-0009-8053-8057} \and \\
Damon Conover\inst{3}\orcidlink{0009-0009-7246-5054} \and 
Guillaume Sartoretti\inst{4}$^\dagger$\orcidlink{0000-0002-7579-9916} \and 
Aniket Bera\inst{1}$^\dagger$\orcidlink{0000-0002-0182-6985}
}

\authorrunning{A.~Vashisth et al.}

\institute{Purdue University, USA \and
Indian Institute of Technology, Kharagpur, India \and
DEVCOM Army Research Lab, USA \and
National University of Singapore, Singapore \\
\email{{vashista}@purdue.edu}}

\maketitle

\begingroup
\renewcommand\thefootnote{}
\footnotetext{\hspace{-0.75em}\textsuperscript{$\ast$} Corresponding Author}
\footnotetext{\hspace{-0.75em}\textsuperscript{\textdagger} These authors contributed equally and share senior authorship.}
\endgroup

\begin{figure*}[t]
\centering
\includegraphics[width=\textwidth]{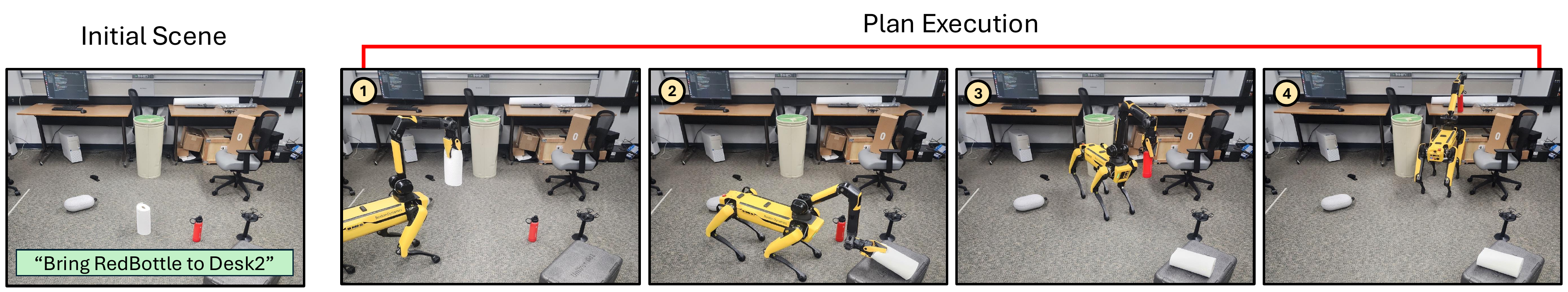}
\caption{Demonstration of our approach, CoReLIN, deployed on the Boston Dynamics Spot robot. CoReLIN first observes the initial scene and determines the presence of task-relevant objects (e.g., Bottle and Desk) alongside environmental clutter. Based on the observations, current task progress, and cost-benefit analysis, our approach determines the best course of action (``Move PaperTowelRoll to BlackBox", ``Move RedBottle to Desk2"). CoReLIN first optimizes the environment by moving the clutter (the paper towel roll) to a carefully chosen out-of-the-way spot (the black box). It can then place the red bottle on the desk.}\small

\label{F:teaser}
\end{figure*}


\begin{abstract}

Robot navigation typically assumes an obstacle-free path exists between start and goal. 
In real environments, however, clutter may block all routes. 
We introduce \textbf{Lifelong Interactive Navigation}, where a mobile robot with manipulation capabilities must move objects to forge paths and complete sequential object-placement tasks. 
Because environment modifications persist, decisions impact future navigability and task difficulty.
We propose \textbf{CoReLIN}, an LLM-driven constraint-based reasoning framework with active perception. 
CoReLIN reasons over a structured scene graph to decide which objects to relocate, where to place them, and where to explore next. 
A standard motion planner executes reliable navigation and manipulation primitives.
To evaluate long-horizon behavior, we introduce $2$ new metrics - \textbf{Long-term Efficiency Score (LES)}, a unified metric capturing success, execution efficiency, environment optimality, captured by \textbf{Price of Clutter}. 
In ProcTHOR-10k, CoReLIN outperforms best baseline by $16\%$ under standard metrics and LES, and transfers to real-world hardware\footnote{Our code and dataset are available at \href{https://accgen99.github.io/projects/corelin/}{project webpage}.}.
  
\end{abstract}

\section{Introduction}
\label{sec:intro}

Visual navigation has achieved remarkable progress in recent years via advances in embodied AI~\citep{kolve2017ai2, anderson2018vision, chaplot2020learning, ahn2022can, huanginner, huang2023voxposer}, generative models~\citep{Zeng2025NaviDiffusorCD, Bar2024NavigationWM}, and deep learning~\citep{Song2024TowardsLV, Mattamala2024WildVN}. However, most navigation systems~\citep{ahn2022can, huanginner} assume the existence of at least one obstacle-free path between the start and goal locations. This assumption holds in idealized cases, but may not hold in real-world scenarios such as homes and warehouses, where clutter, furniture, and everyday objects can obstruct all routes to the goal. In such settings, navigation alone is insufficient and robots must take agency over their environment rather than merely reacting to it. With modern mobile platforms increasingly equipped with arms, grippers, and perception modules, they are no longer just sensors on wheels but embodied agents capable of reshaping their surroundings to achieve their goals. In these cases, an agent must do more than navigate: it must reconfigure its environment to make progress.
 
Existing visual navigation frameworks are typically brittle under such conditions. Reactive planners~\citep{Fox1997TheDW, Likhachev2008PlanningLD} often detour around occlusions, while learned policies~\citep{Mousavian2018VisualRF, wang2023active} fail to generalize to new environments. Interactive navigation methods~\citep{zeng2021pushing, wang2023camp, Wang2024AnIN, du2026fast, zhang2025namo} have begun to address this gap, but are commonly defined for a single task at a time, often pre-arranged such that interaction is the only solution, and assume full observability of the environment. In contrast, real robots face sequential streams of tasks in the same dynamic space, where each decision -- to move or not to move an object -- carries lasting consequences.

In this work, we introduce the \textit{Lifelong Interactive Navigation} problem, where a mobile manipulator receives a sequence of navigation-and-manipulation objectives in an unknown environment, each requiring placement of a given object (eg. Basketball, Alarm Clock) onto a target object (eg. Bed, Arm Chair). This requires the agent to continuously reason about where to explore, which obstacles to remove, which to bypass, and future impact of these choices. 
Crucially, planning and perception are tightly coupled: the agent not only plans about how to act but also where to look next to reveal task-relevant objects. While moving obstacles could open up previously unavailable routes at a nontrivial manipulation cost, detouring may save effort while restricting future mobility. Success thus depends not only on short-term feasibility but also on long-horizon, perception-aware reasoning over the evolving environment.

Large language models (LLMs) can reason over structured and semantic inputs~\citep{ahn2022can, huanginner}, with their strengths demonstrated in abstract reasoning rather than in producing fine-grained control sequences. Our key insight is to invert the typical interface between LLMs and embodied agents: we recast the LLM's role from action sequence generator to a constraint reasoner, operating over the structure of the environment. A structured scene graph, continuously updated with perceptual feedback to include new observations, serves as the LLM’s input representation. The frozen LLM then treats each blocking object as a decision point and determines whether to intervene -- i.e., whether moving the obstacle will substantially expand navigable space given the movement cost -- or if it can be safely detoured around without significant impact on long-term performance. 

By re-framing planning as constraint resolution, we shift long-horizon reasoning from chaining low-level actions into making a small number of strategic, semantically grounded decisions. This enables zero-shot LLM planning, without task-specific fine-tuning, while tightly coupling perception and reasoning.

Building on this constraint-driven formulation, we evaluate our framework in the ProcTHOR-10k~\citep{deitke2022} simulator, which offers visually rich, physics-enabled environments. Our task suite is augmented with cluttered object configurations across environments consisting of up to $10$ rooms. Each episode involves sequential tasks (e.g., bringing object X to receptacle Y), where the robot receives each new task only after completing the previous one. The robot has no prior knowledge of future tasks -- if it fails to complete a task, the episode terminates immediately, and it does not receive any subsequent tasks. Hence, later in the episode, success depends not only on efficient exploration but also on strategic environment modification, as earlier decisions (such as whether to move or detour around obstacles) can significantly influence accessibility to regions containing the target objects, and later progress. By evaluating our system in this online task sequence framework, we test the real-world applicability of our  approach, which must adapt in real time without knowledge of future goals. The key contributions of our work include:
\begin{itemize}
    \item Lifelong Interactive Navigation: Extending prior interactive navigation problem to long-horizon, sequential tasks in unknown environments.

    \item Constraint-based planning framework: Shifting the decision space from low-level actions to high-level environmental constraints, enabling zero-shot long-horizon reasoning by large language models.

    \item Our method achieves the highest Long-term Efficiency Score (LES), improving over the strongest non-learned baseline by $16\%$, and outperforming prior interactive navigation methods by $3-6\times$ in complex environments.
\end{itemize}

\section{Related Work}
\label{sec:rel_work}

\textbf{Embodied and Visual Navigation}. Visual navigation has been extensively studied as mapping observations to actions under the implicit assumption of existence of at least one obstacle-free path between start and goal. Early works and benchmarks such as Cognitive Mapping and Planning~\citep{gupta2017cognitive}, Target-driven Navigation~\citep{zhu2017target}, AI2-THOR~\citep{kolve2017ai2}, Gibson~\citep{Xia2018GibsonER}, Habitat~\citep{savva2019habitat, szot2021habitat}, and ProcTHOR~\citep{deitke2022} have driven progress toward robust policies in visually rich, partially observed environments. Subsequent methods improve exploration and representation via generative models, memory, and structured scene reasoning~\citep{Zeng2025NaviDiffusorCD, Bar2024NavigationWM, Li2024MemoNavWM, Kwon2021VisualGM, Zhang2021HierarchicalOG, Zhai2022PEANUTPA, Singh2023SceneGC, werby2024hierarchical}, and tackle image-goal, language-driven, and multi-object navigation~\citep{Sun2023FGPromptFG, Zhang2024ImagineBG, Wang2024MODDNAC, Yu2024TrajectoryDF, Gadre2022CoWsOP, Marza2022MultiObjectNW}. However, these methods treat the environment as static and ultimately traversable: agents search more efficiently, predict unseen regions, exploit priors over layouts, or decompose the task, but if all geometric routes to the goal are blocked by clutter, they can only attempt detours or fail.  
In contrast, our problem setting explicitly targets cluttered, potentially non-traversable environments, enabling a mobile manipulator to reason about ``whether, which, and where'' to move obstacles, reconfiguring the environment to restore or improve connectivity for long-horizon task sequences. \\

\noindent \textbf{Interactive Navigation $\&$ Navigation Among Movable Obstacles (NA- MO)}. A parallel line of work studies Navigation Among Movable Obstacles, where an agent may relocate objects to reach a blocked goal. Classical NAMO and rearrangement planners search over joint spaces of robot motion and obstacle configurations~\citep{Wilfong1988MotionPI, chen1991practical, Chadzelek1996HeuristicMP, Okada2004EnvironmentMP, stilman2005navigation, stilman2008planning, Nieuwenhuisen2006AnEF, Levihn2012HierarchicalDT}, but assume full knowledge of geometry and dynamics, small number of obstacles, and short horizons, while optimizing reaching a single goal rather than reasoning about persistent environmental changes or future tasks. Interactive navigation approaches bring these ideas into embodied setting: Interactive Gibson~\citep{Xia2018GibsonER} provides benchmarks for interaction in clutter; InterNav~\citep{zeng2021pushing} and CaMP~\citep{wang2023camp} learn how to move blocking objects based on visual input. Interactive-FAR~\citep{He2024InteractiveFARInteractiveFA}, IN-Sight~\citep{Schoch2024INSightIN}, ADIN~\citep{Wang2024AnIN}, SPIN~\citep{Uppal2024SPINSP}, Flax~\citep{du2026fast}, NAMO-LLM~\citep{zhang2025namo} and related systems for legged manipulators or human-in-the-loop assistance explore fast heuristics, affordances, or uncertainty-aware querying for interactive routing. \newstuff{RoboEXP~\citep{jiang2025roboexp} and CuriousBot~\citep{wang2026curiousbot} interact to reveal unknown hidden spaces and discover object relations.}

However, these methods are largely reactive, aiming to locally bypass or momentarily displace obstacles rather than plan how modifications shape future accessibility. NAMO-LLM~\citep{zhang2025namo} does not scale to our highly cluttered scenes (hundreds of obstacles) due to LLM context limits, and FLAX~\citep{du2026fast} targets simple gridworlds with few obstacle instances/types. Moreover, many approaches~\citep{zeng2021pushing, wang2023camp, He2024InteractiveFARInteractiveFA, Schoch2024INSightIN, Wang2024AnIN, Uppal2024SPINSP} are episodic and reset after a single task, avoiding long-horizon manipulation consequences; in contrast, our Lifelong Interactive Navigation runs in a sequential no-reset setting, requiring decisions of which obstacles to move, where to place them, and when to abstain to optimize long-term connectivity and efficiency under uncertainty.

\section{Our Approach}
\begin{figure*}[t]
\centering
\includegraphics[width=\linewidth]{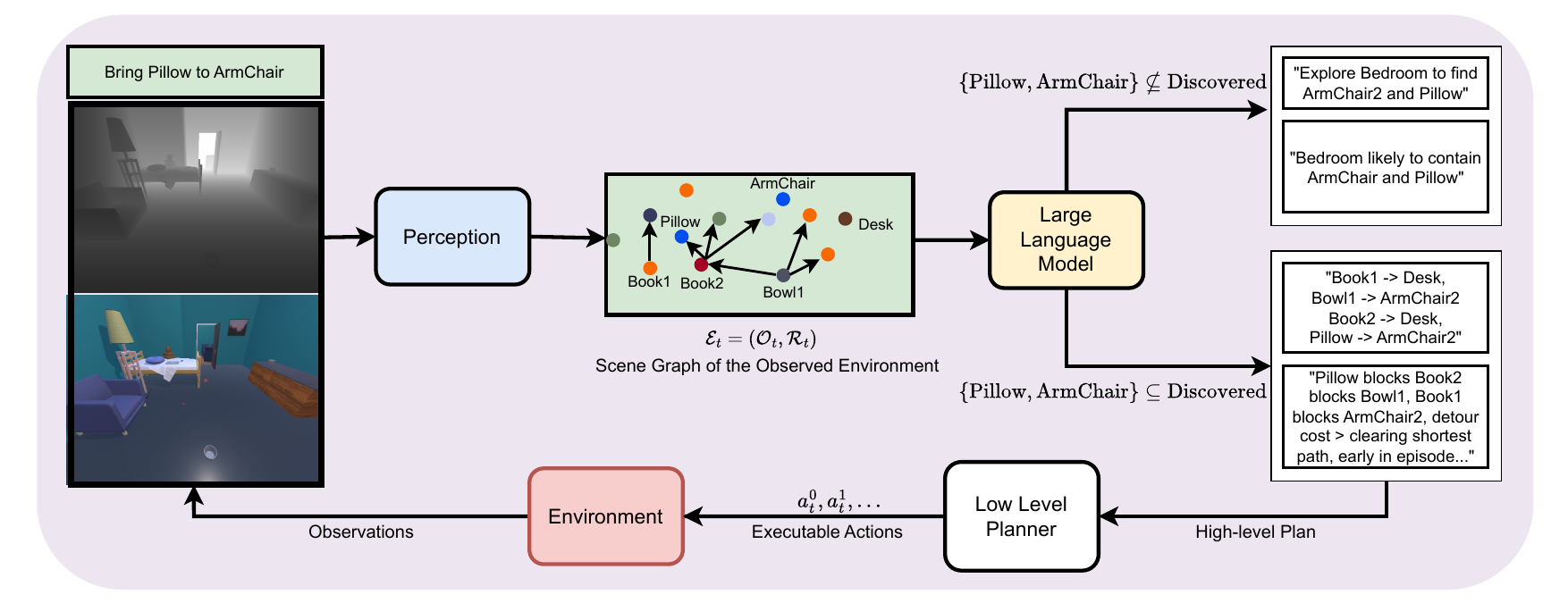}
\caption{Overview of our constraint-based planning framework for interactive navigation. At each timestep, our agent receives observations from the environment, which are utilized by our perception module to update the scene graph. Our scene graph contains the observed objects of the environment as nodes and the blocking relation among them as edges. Each node has its own attributes, which provide additional navigability and manipulability context to the LLM. The LLM then decides whether to explore the environment further to collect additional information or to attempt to complete the given task based on the scene graph content and the environment constraints.}\small

\label{F:overview}
\end{figure*}

\textbf{Overview.}
We target the Lifelong Interactive Navigation setting introduced in Sec.~\ref{sec:intro}, where a mobile manipulator receives a sequence of navigation-and-manipulation tasks in a \newstuff{partially} unknown, cluttered environment. The central challenge is that not all tasks are reachable by pure navigation: some obstacles must be moved, some rooms must be explored, and each intervention can affect future tasks. To handle this, we decouple strategic long-horizon choices from tactical low-level control. 

At time $t$, the agent maintains a partial environment state $\mathcal{E}_t = (\mathcal{O}_t, \mathcal{R}_t)$, where $\mathcal{O}_t$ is the set of discovered objects/rooms and $\mathcal{R}_t$ encodes their blocking and reachability relations. \newstuff{Objects are persistent scene-graph nodes with unique IDs assigned at detection.} Our large language model (LLM) operates on a structured textual serialization of $\mathcal{E}_t$ and reasons about how to resolve the current environmental constraints -- deciding, for example, which obstacle to relocate, where to place it, or which room to investigate next. These high-level constraint resolutions are then converted by a low-level planner into concrete navigation and manipulation actions. \newstuff{CoReLIN operates in a closed-loop \emph{perceive--plan--act} cycle, where each executed action produces new observations that update the scene graph and trigger replanning over the evolving environment.} Hence, we recast the LLM as a constraint reasoner over the environment: it determines ``what to change in the world'' rather than ``how to move the robot step by step''. \\

\noindent \textbf{Perception and structured scene construction.}
In this work, we assume that the general floor plan is known, with room contents unknown at first. We define a grid-graph $G_{t}=(N_{t}, E_{t})$ denoting the discovered and traversable region of the world at time $t$. Here, nodes $N_{t}$ correspond to the nodes that the robot has discovered to be free and can physically occupy, and edges $E_{t}$ connect adjacent free nodes. The graph is constructed incrementally: nodes and edges are added only when free space is observed through the onboard perception system. All path planning and connectivity reasonings are performed over this partial graph.

After each observation, the perception module detects visible objects and free space, updates the grid-graph $G_{t}$ with newly discovered traversable nodes, and augments a structured scene representation that maintains both the set of discovered objects and the set of known but unexplored rooms. Although the total number of rooms and their coarse topological adjacencies are provided by the underlying floorplan, no geometric routes are assumed known a priori. The actual navigable passages between rooms -- including whether doorways, corridors, or connecting regions are free, blocked, or partially occluded -- must be discovered through exploration. As a result, the robot progressively expands the partial graph $G_t$ and discovers the feasible routes between task-relevant objects throughout the task.

We represent the world state at time $t$ as a directed graph $\mathcal{E}_t = (\mathcal{O}_t, \mathcal{R}_t)$, where nodes correspond to discovered objects or rooms, and edges encode blocking relations between them (e.g., an object obstructs the shortest path to another). Formally, for $o_i, o_j \in \mathcal{O}_t$,
\begin{equation}
C_{ij} =
\begin{cases}
1, & \text{if } o_i \text{ blocks the shortest path to } o_j,\\
0, & \text{otherwise.}
\end{cases}
\end{equation}

We utilize the current grid graph $G_{t}$ to describe the impact of an obstacle object on the navigability of the region. Let $n(o_i) \in N_{t}$ denote the node occupied by $o_i$. We compute the (normalized) betweenness centrality of $n(o_i)$:
\[
\mathrm{bc}(n(o_i)) = \frac{1}{Z} \sum_{\substack{s \neq t \in N_{t} \\ s \neq n(o_i) \neq t}}
\frac{|\sigma_{st}(n(o_i))|}{|\sigma_{st}|},
\]
where $|\sigma_{st}|$ is the cardinality of the set of shortest paths $\sigma_{st}$ between any node $s$ and any node $t$ where $s\ne t$, $|\sigma_{st}(n(o_i))|$ is the cardinality of the subset of the shortest paths that pass through $n(o_i)$, $\sigma_{st}(n(o_i))$, and $Z$ is a normalization constant. This measures how many shortest paths depend on the node occupied by $o_i$, and thus how many paths would potentially become available if $o_i$ were removed.

In our representation $\mathcal{E}_t$, each object node $o_i \in \mathcal{O}_t$ is augmented with attributes that capture both geometric and topological context for decision-making:
(i) the cost of traversing the shortest path to the object, if such a path exists;
(ii) the set of discovered objects that block this path, represented via $C_{ij}=1$;
(iii) the betweenness centrality $\mathrm{bc}(n(o_i))$ of the grid cell occupied by the object in the grid graph $G_{t} = (N_{t}, E_{t})$, measuring its influence on global connectivity at time $t$;
(iv) the cost of an alternative path that avoids all current blockers, if such a path exists.
All path computations are performed on $G_{t}$, where only nodes and edges corresponding to currently observed free space are treated as traversable; unknown regions are never assumed passable until they are discovered. Unseen rooms are maintained as frontier nodes, serving as exploration candidates. After each executed action, $\mathcal{E}_t$ is updated to incorporate newly discovered objects, newly navigable rooms, revised exploration status, and obstacles that have been successfully removed. \\

\noindent \textbf{LLM as a constraint-based planner.}
Given this text description of the scene graph, we use a frozen LLM to select the next high-level step. Intuitively, every blocking object is a decision point: should this object be relocated now, and if so, where, or can it be bypassed without significantly harming long-term performance? We formalize this as a cost-benefit selection over blocking objects.

For a candidate obstacle $o_i$, we define a removal cost that accounts for both the effort of manipulating the object and the effort of relocating it to an available location. To ensure they can be appropriately compared, both navigation and manipulation costs are estimated in terms of time:
\[
\mathrm{cost}(o_i, r_t, z_j) = d(r_t,o_i) + 2\times e(o_i) + d(o_i, z_j),
\]
where $d:N \times N \rightarrow \mathbb{R}^+$ describes the time taken to traverse the geodesic distance between two nodes. Hence, $d(r_t,o_i)$ is the cost of navigating from the robot's current position to the obstacle, $e(o_i)$ summarizes the manipulation effort for object $o_i$ (time required to pick and place an object due to its physical attributes such as size and weight), and $d(o_i, z_j)$ is the cost of navigating the path from the obstacle’s position to an available location $z_j$. The values of $d(r_t, o_i)$ and $d(o_i, z_j)$ are estimated from the paths on the known partial grid graph $G_t$. In our experiments, we simplify the manipulation cost estimation using a static value of $e(o_i)=5.00$ (seconds) (counted twice for pick-and-place actions). This setting reflects our empirical observation that a pick/place cycle -- including approach, alignment, grasp execution, verification, and safety pauses -- dominates base travel time, and therefore the cost function should generally favor clearing tasks via short navigation movements over initiating unnecessary manipulations. The LLM reasons jointly over candidate pairs $(o_i, z_j)$, evaluating both the effort required to clear the object and the suitability of each placement (whether the placement location is accessible, obstacles on the path, their attributes and their valid placement locations, etc.). We also present an ablation that varies the value of $e$ to measure the sensitivity of our approach to this hyperparameter.

We intentionally simplify the manipulation aspect of our pipeline: the primary contribution of this work lies in high-level planning and task sequencing and not low-level affordance estimation. Under this assumption, a fixed effort term provides a reasonable approximation of the pick–place cycle duration.
Conceptually, the decision faced by the planner at time $t$ can be viewed as evaluating the trade-off

\begin{equation}
o^\star, z^\star = \argmin_{o_i \in \mathcal{O}_t,\, z_j \in \mathcal{Z}_t} 
\big( \mathrm{cost}(o_i, r_t, z_j) - \, \mathrm{bc}(n(o_i)) \big),
\label{e:1}
\end{equation}
Importantly, \cref{e:1} is not explicitly optimized by our system. Rather, it serves as an analytical abstraction to illustrate the structure and scale of the underlying decision space. In cluttered environments, the planner must implicitly reason over all candidate obstacle–drop-zone pairs $(o_i, z_j)$, a set that grows combinatorially with the number of movable objects and feasible placements.  
\footnote{This formulation is merely intended to highlight the competing factors involved in the task -- manipulation cost, travel distance, and long-term navigability -- and is not utilized as an explicit heuristic during planning.}

When no obstacle offers a sufficient gain, the planner first queries the current grid graph $G_{t}$ to determine whether an alternate, obstacle-free path to the goal exists. If such a detour exists, the agent proceeds with a standard navigation sequence using the low-level planner, completing the task without manipulating any obstacles. If a detour does not exist, the planner attempts to move the minimum number of obstacles that would restore connectivity to achieve the task. If the goal objects have not been discovered yet, the planner initiates a task-directed exploration. \newstuff{Exploration is not a hard-coded perception routine: based on the current task and observed scene graph, the planner explicitly selects a room to explore to gather missing information from, then replans from the updated scene graph.} During exploration, the agent ranks known but unexplored (or partially explored) rooms by semantic relevance to the task. For example, for the task ``bring vase to dining table'', rooms such as Living Room or Kitchen are prioritized over Bathroom or Bedroom. Semantic relevance serves only as a soft ordering, not a hard constraint. If the target objects are not found in higher-ranked rooms, the planner continues exploring remaining rooms in decreasing order of relevance, effectively performing exhaustive search over all reachable rooms. Once the relevant objects are discovered, the planner evaluates connectivity, identifies blocking obstacles, and decides which object to move and where based on obstacle attributes, drop-zone feasibility, and manipulation cost. If all reachable rooms are explored without finding the required objects, the episode ends with failure.

Deciding whether to move an obstacle depends on several factors such as remaining future tasks, long-term cost of detouring around the obstacle (if not moved), immediate cost of moving, and contextual constraints such as available locations for placement and unexplored regions. Hence, in our framework, the LLM operates as a constraint reasoner, combining these quantitative estimates with commonsense and task-level reasoning to generate interpretable, semantically grounded plans that generalize beyond the scope of a fixed heuristic. \\

\noindent \textbf{Low-level planning and closed-loop execution.}
The execution of the LLM's output plan is delegated to a standard motion stack that outputs a sequence of executable actions $[a_t^0,\ a_t^1,\ a_t^2, \cdots]$. Prior to querying the LLM, the environment configuration $\mathcal{E}_t$ is filtered through the low-level planners to ensure that only reachable objects, feasible pick-and-place actions, and collision-free navigation targets are exposed in the decision space. This guarantees that the LLM reasons only over executable actions. For navigation, we run a Dijkstra-based planner over the currently discovered and reachable subgraph to obtain collision-free paths to the target objects, obstacle objects, rooms, or placement locations. For manipulation, we treat pick-place as a reliable primitive provided by the underlying robot control stack, and assume that designated objects have sufficient free space for placement. This separation of concerns lets the LLM focus on long-horizon, semantically meaningful choices about which constraints to resolve (which object to move, where to relocate it, and when to explore), while the low-level planner ensures reliable navigation and manipulation control in cluttered, physics-enabled environments under our assumptions.

\section{Experiments}
\subsection{Setup}
\textbf{Dataset.} We construct a dataset of cluttered indoor floors by instantiating ProcTHOR-$10$k simulator with procedurally generated clutter. We generate $10$k episodes with $20$ tasks each, following the setup described in Sec.~\ref{sec:intro}. For our test set, we randomly draw $100$ environments equally distributed among the number of rooms $1-10$. Following previous works~\citep{wang2023camp}, we group the environments by floorplan complexity and report the results in three categories: $1-3$, $4-6$, and $7-10$. To generate realistic clutter in the environment, we occlude a percentage of traversable nodes (based on floorplan area) in the grid graph $G=(N, E)$ of an environment without obstacles, where the probability of a node $n\in N$ being selected for occlusion is proportional to its betweenness centrality $\mathrm{bc}(n)$. This biases obstacle placement toward structurally critical regions such as hallways, bottlenecks, and intersections, without deterministically blocking all high-centrality nodes. We provide dataset visualization in \cref{app:dataset}. \\

\noindent \textbf{Evaluation Metrics.}
Our framework jointly optimizes three objectives: (i) the success rate $\mathrm{SR} = \frac{1}{N_\text{episodes}} \sum_{k=1}^{N_\text{episodes}} \frac{N_{\text{success}}^{(k)}}{N_{\text{goals}}^{(k)}}$, reflecting the agent's ability to complete assigned tasks (where $N_{\text{success}}^{(k)}$ and $N_{\text{goals}}^{(k)}$ denote the number of successfully completed and total assigned goals in episode $k$, respectively.); (ii) the time efficiency, measured by the number of timesteps required to finish each episode $\mathrm{TS} = \frac{1}{N_\text{episodes}} \sum_{k=1}^{N_\text{episodes}} T_k,$ (where $T_k$ is the total number of timesteps in episode $k$); and (iii) the environmental efficiency for navigation, which quantifies how the agent's actions improve or degrade the navigability of the environment. As explained in~\cref{sec:PoC}, we quantify this via the ratio of the cumulative shortest-path lengths in the current cluttered environment to those in an obstacle-free configuration, termed as \textbf{Price of Clutter} - $\mathrm{PoC} = \frac{\sum_{s,t \in V} d_{\text{obs}}(s,t)}{\sum_{s,t \in V} d_{\text{free}}(s,t)},$ where $d_{\text{obs}}(s,t)$ is the shortest-path distance between nodes $s$ and $t$ in the cluttered environment, and $d_{\text{free}}(s,t)$ is the corresponding distance in the obstacle-free floorplan, both evaluated over the complete graph $G$.

In our problem setting, SR captures the task completion, TS measures execution speed, and PoC measuring lifelong environment that remains navigable for future goals. To consolidate these competing objectives into a single primary metric, we introduce the globally normalized \textbf{Long-term Efficiency Score (LES)}, which models the trade-off between short-term success (SR), temporal efficiency (TS), and sustainable environment restructuring (PoC). Concretely, we first convert TS and PoC into higher-is-better utilities using global min-max normalization computed over the evaluation set:
\begin{equation}
u_{\text{TS}} = \!1 - \frac{\text{TS}-\text{TS}_{\min}}{\text{TS}_{\max}-\text{TS}_{\min}},\quad
u_{\text{PoC}} = \!1 - \frac{\text{PoC}-\text{PoC}_{\min}}{\text{PoC}_{\max}-\text{PoC}_{\min}},
\end{equation}
where $\text{TS}_{\min}, \text{TS}_{\max}, \text{PoC}_{\min}, \text{PoC}_{\max}$ are global extrema. We then aggregate the three objectives using a weighted geometric mean:
\begin{equation}
\text{LES} = 100 \cdot \text{SR}^{w_{\text{SR}}}\cdot (u_{\text{TS}}+\epsilon)^{w_{\text{TS}}}\cdot (u_{\text{PoC}}+\epsilon)^{w_{\text{PoC}}},
\qquad
w_{\text{SR}}+w_{\text{TS}}+w_{\text{PoC}}=1,
\end{equation}
with $w_{\text{SR}}=0.5$, $w_{\text{PoC}}=0.25$, $w_{\text{TS}}=0.25$, and $\epsilon=10^{-8}$ for numerical stability.
A high LES indicates that a policy not only succeeds, but does so efficiently while preserving long-term navigability. By construction, higher SR increases LES, whereas larger TS or PoC decreases LES (via lower utilities), and the multiplicative form prevents a method from appearing strong by excelling in only one dimension (e.g., being fast but failing often, or succeeding while severely degrading the environment). We extensively describe our metric LES in \cref{sec:LES}.

\subsection{Baseline Comparison}

\begin{table*}[h]
\centering
\resizebox{\textwidth}{!}{%
\begin{tabular}{l|c c c c|c c c c|c c c c}
\toprule

\textbf{Episode Horizon} & 
\multicolumn{4}{c|}{\textbf{$\mathbf{1-3}$ rooms}} & 
\multicolumn{4}{c|}{$\mathbf{4-6}$\textbf{ rooms}} & 
\multicolumn{4}{c}{$\mathbf{7-10}$\textbf{ rooms}} \\
\cmidrule(lr){2-5} \cmidrule(lr){6-9} \cmidrule(lr){10-13}
 & SR & PoC & TS & LES
 & SR & PoC & TS & LES
 & SR & PoC & TS & LES\\

\midrule
$\mathbf{g=10}$ \\
CoReLIN (known) & $98.79$ & $1.29$ & $578.88$ & $\underline{94.52}$ & $100.00$ & $1.28$ & $1040.42$ & $\mathbf{91.60}$ & $90.61$ & $1.54$ & $1594.36$ & $\mathbf{81.62}$\\

CoReLIN (unk) & $93.33$ & $1.85$ & $675.52$ & $86.30$ & $72.42$ & $2.12$ & $2099.79$ & $65.64$ & $65.45$ & $3.00$ & $2270.42$ & $54.14$\\

ADIN & $8.57$ & $1.81$ & $419.64$ & $26.90$ & $3.93$ & $1.44$ & $955.25$ & $17.91$ & $3.21$ & $2.90$ & $1462.14$ & $13.30$\\

InterNav & $1.64$ & $1.90$ & $496.09$ & $37.30$ & $0.88$ & $1.49$ & $2155.88$ & $28.44$ & $0.52$ & $1.43$ & $4511.78$ & $21.71$\\

FastDownward & $38.12$ & $2.47$ & $87.50$ & $53.28$ & $17.50$ & $2.41$ & $67.94$ & $36.43$ & $7.74$ & $3.14$ & $30.42$ & $21.45$\\

Always Interact & $100.00$ & $1.19$ & $681.79$ & $\mathbf{94.60}$ & $100.00$ & $1.37$ & $1521.58$ & $\underline{87.57}$ & $100.00$ & $1.16$ & $2565.52$ & $\underline{79.67}$\\

Always Detour & $87.88$ & $2.34$ & $364.48$ & $81.00$ & $86.36$ & $2.10$ & $613.52$ & $81.24$ & $66.06$ & $3.23$ & $758.76$ & $59.39$\\

Clean + S/P & $100.00$ & $1.00$ & $835.72$ & $94.48$ & $100.00$ & $1.00$ & $1909.91$ & $86.60$ & $100.00$ & $1.00$ & $3865.10$ & $57.91$\\

\midrule

$\mathbf{g=15}$ \\
CoReLIN (known) & $100.00$ & $1.22$ & $814.48$ & $\underline{94.35}$ & $97.78$ & $1.20$ & $1403.76$ & $\mathbf{90.18}$ & $94.14$ & $1.57$ & $2356.94$ & $\mathbf{79.90}$\\

CoReLIN (unk) & $94.14$ & $1.53$ & $857.79$ & $89.02$ & $79.41$ & $1.46$ & $2049.79$ & $76.17$ & $64.85$ & $2.88$ & $2843.94$ & $54.29$\\

ADIN & $13.81$ & $1.81$ & $602.86$ & $8.77$ & $4.76$ & $1.43$ & $1453.61$ & $5.11$ & $2.86$ & $2.90$ & $2178.04$ & $3.12$\\

InterNav & $17.75$ & $1.81$ & $688.76$ & $39.91$ & $12.62$ & $1.51$ & $2950.67$ & $34.36$ & $4.17$ & $1.43$ & $6545.78$ & $19.52$\\

FastDownward & $31.67$ & $2.33$ & $104.41$ & $49.37$ & $12.53$ & $2.38$ & $69.24$ & $31.01$ & $5.42$ & $3.28$ & $34.00$ & $17.57$\\

Always Interact & $100.00$ & $1.15$ & $905.39$ & $\mathbf{94.43}$ & $100.00$ & $1.34$ & $1983.30$ & $\underline{86.56}$ & $100.00$ & $1.15$ & $3321.25$ & $\underline{77.16}$\\

Always Detour & $84.04$ & $2.19$ & $505.30$ & $80.37$ & $84.44$ & $2.09$ & $878.97$ & $79.80$ & $66.26$ & $3.21$ & $1094.18$ & $59.16$\\

Clean + S/P & $100.00$ & $1.00$ & $1075.30$ & $92.45$ & $100.00$ & $1.00$ & $2383.58$ & $85.78$ & $100.00$ & $1.00$ & $4693.99$ & $56.49$\\

\midrule 
$\mathbf{g=20}$ \\
CoReLIN (known) & $94.55$ & $1.50$ & $1044.55$ & $89.20$ & $97.73$ & $1.25$ & $1911.97$ & $\mathbf{88.37}$ & $100.00$ & $1.23$ & $2435.48$ & $\mathbf{86.56}$\\ 

CoReLIN (unk) & $88.94$ & $1.48$ & $1339.36$ & $83.87$ & $81.67$ & $1.81$ & $3211.20$ & $67.25$ & $62.27$ & $2.79$ & $4497.80$ & $49.99$\\ 

ADIN & $13.81$ & $1.81$ & $602.86$ & $37.42$ & $4.76$ & $1.43$ & $1453.61$ & $20.03$ & $2.86$ & $2.90$ & $2178.04$ & $12.21$\\

InterNav & $16.67$ & $1.96$ & $842.00$ & $33.78$ & $6.00$ & $2.13$ & $4700.70$ & $14.53$ & $3.00$ & $3.34$ & $6089.91$ & $0.20$\\ 

FastDownward & $27.66$ & $2.32$ & $123.19$ & $46.14$ & $9.85$ & $2.38$ & $73.12$ & $27.50$ & $4.06$ & $3.28$ & $35.81$ & $15.21$\\

Always Interact & $100.00$ & $1.14$ & $1108.88$ & $\underline{94.18}$ & $100.0$ & $1.13$ & $2416.67$ & $\underline{87.19}$ & $100.00$ & $1.36$ & $3974.88$ & $\underline{74.73}$\\ 

Always Detour & $90.45$ & $1.96$ & $665.64$ & $85.12$ & $83.18$ & $2.15$ & $1115.27$ & $78.26$ & $64.09$ & $3.43$ & $1381.61$ & $54.99$\\ 

Clean + S/P & $100.00$ & $1.00$ & $1200.48$ & $\mathbf{94.38}$ & $100.00$ & $1.00$ & $2622.94$ & $86.65$ & $100.00$ & $1.00$ & $5169.58$ & $61.86$\\ 

\bottomrule
\end{tabular}
}
\caption{Baseline comparison of our approach CoReLIN with other methods across different floorplans (rooms $1$ through $10$) and varying episode horizons $g\in\{10,15,20\}$ (where $g$ is the total number of tasks per episode). Bold indicates best performing method and underline indicates second-best performing approach.}\small 
\label{table:base_comp}
\end{table*}

We compare our method against $4$ heuristic methods and $2$ learning baselines, using identical goal sequences and obstacle placements across $3$ goal horizons - $g\in\{10, 15, 20\}$. Since the $6$ baselines assume ground-truth knowledge of the environment, we expose this information to one of our variants and report metrics for two versions of our approach - Ours(unk), which has no a priori knowledge of the world and must explore the environment, and Ours (known), which has ground-truth knowledge of the world. Our baselines are: (a)~\textbf{ADIN} (learning-based)~\citep{Wang2024AnIN} (b)~\textbf{InterNav (learning-based)}~\citep{zeng2021pushing} (c)~\textbf{FastDownward} (plans on the complete PDDL problem instance)~\citep{helmert2006fast} (d)~\textbf{Always Detour} (pure navigation) (e)~\textbf{Always Interact} (relocate obstacles on shortest path to the goal) (f)~\textbf{Clean + S/P} (First relocate all obstacles in the environment, then perform pure navigation). We adapt and re-train the learning baselines on our dataset.

Across medium and large environments ($4-6$ and $7-10$ rooms), our approach achieves the best LES for every goal horizon, and the margin grows as scenes and horizons become complex. 
\newstuff{This is where lifelong structure matters: Always Interact myopically clears each current path, accumulating unnecessary relocations and future traversal cost; Always Detour avoids manipulation but preserves bottlenecks and fails when no free route exists; finally, Clean+S/P restores global connectivity only by paying a large, task-agnostic clearing cost.}
In contrast, our method explicitly balances task completion, time, and environmental impact, producing plans that remain efficient over long horizons without accumulating unnecessary clutter. 
In small environments ($1-3$ rooms), the best heuristics can marginally edge out our LES because navigation distances are inherently short and clutter has limited long-term impact. Simple strategies that aggressively manipulate obstacles can be ``good enough'' and occasionally slightly faster.

The learning-based baselines ADIN~\citep{Wang2024AnIN} and InterNav~\citep{zeng2021pushing} underperform primarily due to a mismatch between their design assumptions and our lifelong setting. 
Both methods are largely optimized for single-goal interactive navigation, where interactions are chosen to unblock the current path to improve immediate reachability. 
In our problem, however, object placements persist and affect future tasks, causing myopic interactions to accumulate clutter and degrade long-term navigability (reflected in PoC), sharply lowering LES. 
As environments and horizons grow, this lack of explicit long-horizon reasoning causes their performance to degrade, whereas our method optimizes decisions for sustained efficiency across the entire sequence.
These approaches were not designed to optimize a global, horizon-spanning objective that explicitly penalizes accumulated clutter, hence they struggle as soon as the sequential goals make ``interaction for the current step'' disadvantageous. 
Notably, our approach Ours (unk) remains competitive and consistently improves over the learning-based baselines, indicating that the gains in larger environments arise from better long-horizon decision-making under exploration rather than privileged access to ground truth.
Hence, our constraint-based planner CoReLIN emerges as the most effective strategy for lifelong interactive navigation. 

We provide the implementation details for our baselines in \cref{sec:baselines}, and additional performance analyses in \cref{sec:base_comp_ext}. 
Additionally, we evaluate our approach under a range of clutter generation policies (\cref{sec:clutter_gen_policy}) and observe consistent trends: our method remains robust and continues to outperform the baselines across policies.

\begin{table*}[t]
\centering
\resizebox{\textwidth}{!}{%
\begin{tabular}{l|c c c c|c c c c|c c c c}
\toprule

\textbf{Manipulation Effort} & 
\multicolumn{4}{c|}{\textbf{$\mathbf{1-3}$ rooms}} & 
\multicolumn{4}{c|}{$\mathbf{4-6}$\textbf{ rooms}} & 
\multicolumn{4}{c}{$\mathbf{7-10}$\textbf{ rooms}} \\
\cmidrule(lr){2-5} \cmidrule(lr){6-9} \cmidrule(lr){10-13}
 CoReLIN (unk) & SR$\uparrow$ & PoC$\downarrow$ & IE & PL (m)$\downarrow$
 & SR$\uparrow$ & PoC$\downarrow$ & IE & PL (m)$\downarrow$
 & SR$\uparrow$ & PoC$\downarrow$ & IE & PL (m)$\downarrow$\\

\midrule

$e=1$ & $92.58$ & $\mathbf{1.44}$ & $42.21$ & $302.73$ & $\textbf{83.79}$ & $\mathbf{1.46}$ & $41.09$ & $\textbf{646.35}$ & $65.45$ & $\mathbf{2.65}$ & $36.09$ & $\textbf{999.48}$\\ 

$e=5$ & $85.30$ & $1.48$ & $36.31$ & $295.04$ & $79.55$ & $1.81$ & $38.48$ & $825.78$ & $\textbf{69.09}$ & $2.79$ & $32.09$ & $1057.96$\\ 

$e=10$ & $92.42$ & $1.53$ & $23.27$ & $301.07$ & $80.45$ & $1.62$ & $39.76$ & $809.92$ & $60.61$ & $2.94$ & $25.48$ & $1211.33$\\ 

$e=15$ & $91.21$ & $1.59$ & $15.33$ & $287.38$ & $80.91$ & $1.87$ & $21.88$ & $822.90$ & $63.39$ & $3.00$ & $20.89$ & $1037.37$\\ 

$e=20$ & $\textbf{92.88}$ & $1.62$ & $9.58$ & $\textbf{259.63}$ & $78.48$ & $1.90$ & $21.36$ & $715.82$ & $60.30$ & $2.95$ & $18.58$ & $1078.59$\\ 

\bottomrule
\end{tabular}
}
\caption{Effect of varying manipulation effort $e$ on the behavior of our approach CoReLIN. $\downarrow$ indicates lower value is better, and $\uparrow$ indicates higher value is better. Bold indicates the best performance of the metric.}\small 
\label{tab:abl_cost}
\end{table*}

\subsection{Analyses}

\textbf{a. Effect of Manipulation Cost:} To examine the behavior of our planner as the cost of manipulating objects $e$ increases, we systematically vary the per-object manipulation cost $e\in[1, 5, 10, 15, 20]$ and measure SR, PoC, and total distance traversed by the robot (PL) across the test set (\cref{tab:abl_cost}).
We report an additional metric of Interaction Encounters (IE) that describes the percentage of obstacles the agent chose to move during the episode.
As the value of $e$ increases, IE decreases sharply across all settings, approximately $4\times$ between $e=1$ and $e=20$, demonstrating that the agent actively avoids unnecessary manipulation when it is costly to do so.
Correspondingly, PoC rises, indicating that the planner tolerates higher clutter levels instead of restoring full connectivity at excessive expense. 
Despite this trade-off, SR remains largely stable, showing that the LLM-driven reasoning identifies and removes only critical obstacles essential for task completion.
Path length (PL) also either remains bounded or slightly decreases, suggesting that when manipulation is de-incentivized, the agent effectively compensates by discovering alternate routes rather than interacting with high-cost obstacles.
Overall, our results confirm that our constraint-based planner internalizes manipulation cost and dynamically strategizes by favoring selective, high-impact interactions over exhaustive environment cleanup, without significant degradation in SR. \\

\noindent \textbf{b. History Length:} We analyze how the amount of prior context $h$ affects performance using LES as the primary metric (\cref{tab:abl_hlen}). In $1-3$ rooms, LES increases monotonically with $h$, reflecting slightly higher SR and markedly lower TS as the planner reuses past decisions and avoids redundant exploration/manipulation. In $4-6$ rooms, LES improves substantially with longer histories. Although SR fluctuates, both TS and PoC benefit from better global consistency -- longer context helps the LLM recall earlier constraints and target fewer, more central obstacles. In $7-10$ rooms, $h=3$ yields the best LES value; the gain at $h=3$ is contributed by an improvement in SR (while maintaining comparable TS and PoC), indicating that extended context accelerates progress once the agent has accumulated a rich constraint history. Overall, longer histories improve or match the best LES across floorplans, as additional context improves constraint reasoning: the planner avoids revisiting explored frontiers, concentrates interactions on high-impact bottlenecks, and reduces detours—thereby balancing success, execution time, and long-term environment optimality. \\

\begin{table*}[t]
\centering
\resizebox{\textwidth}{!}{%
\begin{tabular}{l|c c c c|c c c c|c c c c}
\toprule

\textbf{History Length} & 
\multicolumn{4}{c|}{\textbf{$\mathbf{1-3}$ rooms}} & 
\multicolumn{4}{c|}{$\mathbf{4-6}$\textbf{ rooms}} & 
\multicolumn{4}{c}{$\mathbf{7-10}$\textbf{ rooms}} \\
\cmidrule(lr){2-5} \cmidrule(lr){6-9} \cmidrule(lr){10-13}
CoReLIN (unk) & SR$\uparrow$ & PoC$\downarrow$ & TS$\downarrow$ & LES$\uparrow$
 & SR$\uparrow$ & PoC$\downarrow$ & TS$\downarrow$ & LES$\uparrow$
 & SR$\uparrow$ & PoC$\downarrow$ & TS$\downarrow$ & LES$\uparrow$\\

\midrule
$h=1$ & $89.85$ & $\textbf{1.44}$ & $1793.97$ & $86.50$ & $\textbf{78.94}$ & $1.67$ & $3516.03$ & $71.84$ & $55.00$ & $\textbf{2.71}$ & $5641.48$ & $41.37$ \\ 

$h=3$ & $88.18$ & $1.49$ & $1492.15$ & $86.46$ & $71.36$ & $1.73$ & $2926.48$ & $70.42$ & $\textbf{64.59}$ & $2.78$ & $3334.70$ & $\mathbf{52.60}$ \\ 

$h=6$ & $91.88$ & $1.52$ & $1444.21$ & $88.64$ & $77.12$ & $1.80$ & $3175.39$ & $71.54$ & $57.43$ & $2.91$ & $3996.73$ & $49.67$ \\ 

$h=9$ & $\textbf{92.88}$ & $1.52$ & $\textbf{1403.88}$ & $\mathbf{88.77}$ & $75.40$ & $\textbf{1.45}$ & $\textbf{2634.03}$ & $\mathbf{75.82}$ & $53.24$ & $2.79$ & $\textbf{1228.72}$ & $49.66$ \\ 

\bottomrule
\end{tabular}
}
\caption{Effect of varying historical context length $h$ on the behavior of our approach. $\uparrow$ indicates higher value is better for the metric. Bold indicates the best performance.}\small
\label{tab:abl_hlen}
\end{table*}

\begin{table*}[t]
\centering
\resizebox{\textwidth}{!}{%
\begin{tabular}{l|c c c c|c c c c|c c c c}
\toprule

\textbf{LLM Name} & 
\multicolumn{4}{c|}{\textbf{$\mathbf{1-3}$ rooms}} & 
\multicolumn{4}{c|}{$\mathbf{4-6}$\textbf{ rooms}} & 
\multicolumn{4}{c}{$\mathbf{7-10}$\textbf{ rooms}} \\
\cmidrule(lr){2-5} \cmidrule(lr){6-9} \cmidrule(lr){10-13}
 & SR$\uparrow$ & PoC$\downarrow$ & TS$\downarrow$ & LES$\uparrow$
 & SR$\uparrow$ & PoC$\downarrow$ & TS$\downarrow$ & LES$\uparrow$
 & SR$\uparrow$ & PoC$\downarrow$ & TS$\downarrow$ & LES$\uparrow$\\

\midrule

Gemini & $\underline{87.12}$ & $1.79$ & $2094.79$ & $\underline{79.14}$ & $\textbf{84.09}$ & $\textbf{1.67}$ & $\underline{3545.79}$ & $\mathbf{68.44}$ & $\textbf{71.06}$ & $\textbf{2.73}$ & $4768.48$ & $\mathbf{42.53}$\\ 

GPT-5 & $\textbf{92.27}$ & $\textbf{1.44}$ & $\underline{1732.55}$ & $\mathbf{86.49}$ & $\underline{59.05}$ & $\underline{1.70}$ & $4037.52$ & $53.11$ & $\underline{49.39}$ & $\underline{2.87}$ & $\underline{4107.61}$ & $\underline{40.78}$\\ 

Deepseek & $76.21$ & $\underline{1.63}$ & $\textbf{1549.21}$ & $78.12$ & $51.52$ & $1.82$ & $\textbf{2506.70}$ & $\underline{58.54}$ & $38.03$ & $2.94$ & $\textbf{3305.52}$ & $39.56$\\ 

\bottomrule
\end{tabular}
}
\caption{Performance of different large language models in our framework. $\uparrow$ indicates higher value is better for the metric. Bold indicates the best performance while underline indicates the second-best performance.\small}
\label{tab:abl_llms}
\end{table*}

\begin{table*}[t]
\centering
\resizebox{\textwidth}{!}{%
\begin{tabular}{l|c c c c|c c c c|c c c c}
\toprule

\textbf{Obstacle Density} & 
\multicolumn{4}{c|}{\textbf{$\mathbf{1-3}$ rooms}} & 
\multicolumn{4}{c|}{$\mathbf{4-6}$\textbf{ rooms}} & 
\multicolumn{4}{c}{$\mathbf{7-10}$\textbf{ rooms}} \\
\cmidrule(lr){2-5} \cmidrule(lr){6-9} \cmidrule(lr){10-13}
 & SR & PoC & IE & TS
 & SR & PoC & IE & TS
 & SR & PoC & IE & TS\\

\midrule

$d=0.5$ & $86.52$ & $1.31$ & $39.24$ & $1678.06$ & $80.91$ & $1.26$ & $47.79$ & $3351.85$ & $65.00$ & $2.50$ & $47.39$ & $4748.91$\\ 

$d=1.0$ & $88.94$ & $1.48$ & $39.41$ & $1339.36$ & $81.67$ & $1.81$ & $39.79$ & $3211.20$ & $62.27$ & $2.79$ & $32.70$ & $4497.80$\\ 

$d=2.0$ & $81.88$ & $3.19$ & $16.06$ & $1592.34$ & $55.15$ & $3.67$ & $27.39$ & $3135.45$ & $48.79$ & $4.55$ & $23.21$ & $4773.21$\\ 

\bottomrule
\end{tabular}
}
\caption{Performance of our approach as the obstacle density varies across floorplans.\small}
\label{tab:abl_dense}
\end{table*}

\noindent \textbf{c. Effect of Language Model Choice:} We further evaluate the impact of different large language models within our framework in \cref{tab:abl_llms}. All LLMs are queried with the same prompt (described in ~\cref{sec:LLM_prompts}) and structured scene-graph serialization, hence the differences primarily reflect model-dependent decision behavior. In $1-3$ rooms, \textbf{GPT-5} achieves the best LES via the highest SR, which we attribute to locally consistent choices that work well when the reasoning horizon is short. However, GPT-5 degrades sharply as complexity grows, suggesting brittleness under long-horizon coupling: it tends to over-commit to salient local obstacle relations and under-weight global connectivity and the downstream effects of persistent rearrangements, leading to more harmful placements and reduced SR. \textbf{Gemini} is most robust in larger environments, achieving the best LES in $4-6$ and $7-10$ rooms while maintaining the highest SR. This suggests it better internalizes the ``persistent world'' constraint, favoring interventions that preserve broader navigability and avoid cascading clutter. \textbf{DeepSeek} is less consistent overall, likely due to noisier interpretation of the structured serialization, but it surpasses GPT-5 in the $4-6$ room setting, indicating that failures are driven by planning priors rather than raw language ability. \\

\noindent \textbf{d. Obstacle Density:} 
We assess the robustness of CoReLIN as the obstacle density $d$ increases from $0.5$ (sparse) to $2.0$ (dense), keeping task configurations fixed (\cref{tab:abl_dense}). 
Overall performance declines with higher $d$, as reduced free space hinders long-horizon planning. 
In $1-3$ rooms, SR drops moderately, while PoC more than doubles, indicating diminished environment optimality. 
TS shows non-monotonic behavior as dense clutter sometimes shortens exploration but also causes premature failures. 
This trend intensifies in larger environments, where SR degrades sharply and PoC rises $2\times$ to $3\times$, reflecting that, while the agent must detour or manipulate more often, it struggles to restore global connectivity. 
The decline in interaction encounters (IE) suggests increased conservative approach: as clutter grows, the agent avoids manipulations with low expected payoff. 
Together, these results highlight that higher obstacle density compounds perception and reasoning difficulty, limiting the agent’s ability to maintain long-term environmental optimality. 
We provide analysis of performance comparison of our approach with baselines across varying obstacle density in \cref{sec:dense_perf}. \\

Additionally, we present a detailed failure-case analysis of CoReLIN in \cref{sec:failure_analysis}, 
along with a step-by-step run-through in \cref{sec:run_through}.

\subsection{Hardware Experiments}
We validate sim-to-real transfer by deploying our framework on a Boston Dynamics Spot with the Spot Arm (\cref{F:teaser}). Perception uses only the front fisheye cameras, while the wrist camera supports grasping and placement verification. Metric localization comes from the onboard state estimator, and stereo depth with category-level segmentation incrementally updates a structured scene graph in real time. Experiments are conducted in a three-room indoor environment with $\approx 50.0$ m$^2$ navigable space, evaluating tasks similar to simulation and limited to objects within the gripper envelope. The planner interleaves exploration and targeted rearrangement based on incremental scene updates from fisheye RGB-D observations. This setup preserves our lifelong, partially observed, cluttered assumptions without task-specific fine-tuning or hardware-specific prompts. Qualitatively, behavior mirrors simulation: selective removal of central obstacles, efficient detours when beneficial, and robust retrieve-and-place execution under real sensing and actuation noise. Implementation details are provided in \cref{sec:Spot}.

\section{Limitations and Future Work}
Several extensions to CoReLIN remain promising. We currently model manipulation effort with a fixed cost; future work could infer object-specific effort directly from RGB-D observations to enable richer cost-aware reasoning. Our interaction space is limited to pickup-and-place of small obstacles, and extending it to include pushing, whole-body interaction, or larger objects would broaden applicability.
We also decouple high-level constraint reasoning from low-level motion control for modularity; tighter integration could further improve whole-body efficiency. Scaling to extreme clutter and richer partial observability remains challenging, motivating stronger uncertainty-aware perception and structural priors.
Finally, extending our framework to multi-agent collaborative settings -- where multiple robots coordinate exploration and environment restructuring -- offers an exciting direction for future lifelong embodied systems.

\section{Conclusion}
Our approach redefines the role of large language models in embodied decision-making. Specifically, we recast the role of LLMs as constraint reasoners rather than sequence generators. Our method leverages structured scene graphs and cost–benefit abstractions to choose which environmental constraints to resolve, such as which obstacles to relocate where, or which regions to explore, based on their long-term impact on performance. Through extensive experiments, we showcase that our formulation enables zero-shot, long-horizon planning in unknown environments, achieving competitive task success with substantially fewer manipulations and lower cumulative cost compared to fully informed baselines. Finally, we validate the feasibility of our method on a Boston Dynamics Spot robot equipped with a depth camera and manipulator arm, demonstrating effective active perception and environment restructuring.

\section{Acknowledgements}
This work was supported by the DEVCOM Army Research Laboratory under cooperative agreement W911NF2520170.

\bibliographystyle{splncs04}
\bibliography{main}

\newpage 
\appendix
\section*{Appendix}
\renewcommand{\thesection}{\Alph{section}}
\setcounter{section}{0}

\section{Dataset}
\label{app:dataset}

\begin{figure*}[h]
\centering
\includegraphics[width=\linewidth]{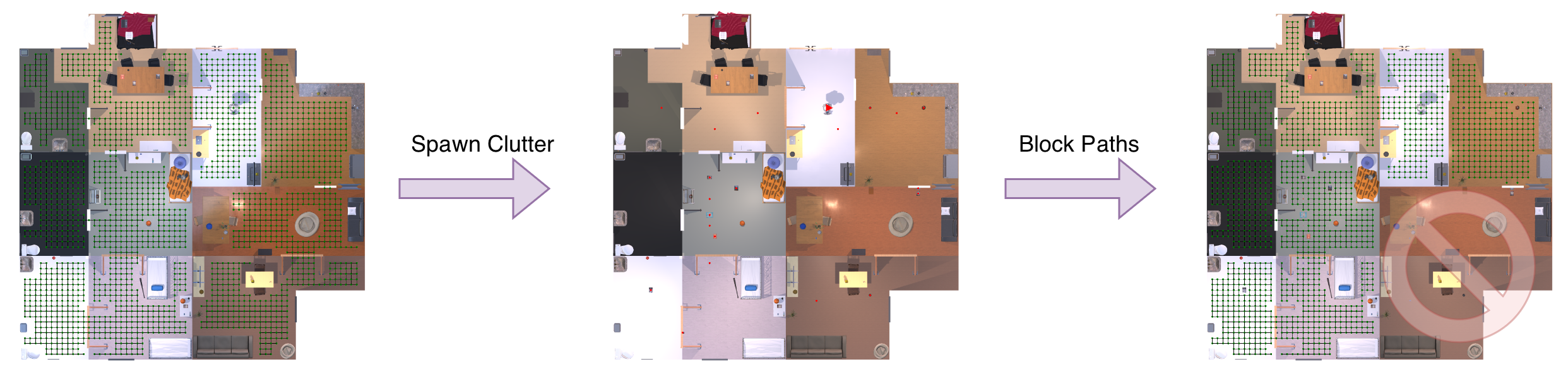}
\caption{Example dataset generation process. In this instance, we consider a $10$ room floorplan from ProcTHOR-$10$k. The left image shows the ideal environment without clutter and completely free for optimal navigation throughout the environment. We then generate clutter in the free space using the between-ness centrality of nodes in $G_{\text{free}}$. The generated obstacle locations are indicated by red circles in the middle image. This causes the blocking of the paths within the accessible rooms, as well as eliminating access to the two rooms on bottom-right of the map, as shown in the right image.}\small

\label{F:app_data_gen}
\end{figure*}

We construct a new dataset of cluttered indoor floors by instantiating ProcTHOR -10k simulator with procedurally generated clutter. As training set for learning baselines, we generate $10$k episodes with $20$ tasks each, following the Lifelong Interactive Navigation setup described in Sec.~\ref{sec:intro}. 
Our test set consists of $100$ environments equally distributed among the number of rooms $1-10$. 

To generate realistic clutter in the environment, we occlude a percentage of traversable nodes (based on floorplan area) in the grid graph $G=(N, E)$ of the unoccluded environment, where the probability of a node $n\in N$ being selected for occlusion is proportional to its betweenness centrality $\mathrm{bc}(n)$. This biases obstacle placement toward structurally critical regions such as hallways, bottlenecks, and intersections, without deterministically blocking all high-centrality nodes. \cref{F:app_data_gen} shows an instance of the process of dataset generation. \cref{F:app_data_viz} shows examples of the generated floorplans.

\section{Price of Clutter}
\label{sec:PoC}
In lifelong interactive navigation, an agent’s actions modify the physical structure of its environment, thereby influencing its future navigability. 
In single-goal episodes, where each episode involves completing only one task, the placement of a moved obstacle has no long-term consequence on performance. 
Hence, such settings offer little incentive for the agent to reason about `where' to relocate an obstacle once removed. In contrast, long-horizon, multi-task episodes demand careful reasoning about object placement. 
An object relocated to a frequently traversed region can block critical routes, forcing repeated manipulations in later tasks. 
Existing interactive navigation approaches, designed for single-task settings, neither quantify such long-term environmental degradation nor explicitly optimize against it. 
To address this, we introduce the \textbf{Price of Clutter (PoC)} metric, which quantifies how much the environment deviates from its obstacle-free optimal configuration. 
Formally, we define:
\begin{equation*}
    \mathrm{PoC} = 
    \frac{\sum_{s,t \in N} d_{\text{obs}}(s,t)}
    {\sum_{s,t \in N} d_{\text{free}}(s,t)},
\end{equation*}
where $d_{\text{obs}}(s,t)$ denotes the shortest-path length between nodes $s$ and $t$ in the cluttered environment, and $d_{\text{free}}(s,t)$ is the corresponding path length in the obstacle-free environment. \\

\noindent \textbf{Calculation:} We consider two versions of the same environment $\mathcal{E}$: 
(i) a fully traversable configuration $\mathcal{E}_{\text{free}}$ with navigable graph $G_{\text{free}}=(N,E)$, and 
(ii) a partially occluded configuration $\mathcal{E}_{\text{curr}}$, where obstacles block certain regions, yielding a reduced graph $G_{\text{curr}}=(N',E')$. 
Nodes present in $G_{\text{free}}$ may be occluded and therefore absent in $G_{\text{curr}}$. 
If either the source $s$ or target $t$ of a shortest-path pair in $G_{\text{free}}$ becomes occluded, that pair is excluded from the computation since it is no longer traversable. 
Such occlusions indirectly increase PoC: when key nodes are removed, surviving paths between other pairs must detour around blocked regions, increasing total path lengths.
For paths that originate or terminate at one of the nodes available in $G_{\text{free}}$ but occluded in $G_{\text{curr}}$, the path is not traversable, and hence the length is infinite.
To keep the metric finite and comparable across scales, we replace infinite distances with a capped maximum equal to $10\times$ the maximum finite shortest-path length observed among surviving nodes. 
This prevents divergence in heavily occluded scenes while still penalizing disconnections proportionally to environment size. \\

\noindent \textbf{Interpretation:} A lower PoC indicates a more globally navigable and optimally structured environment, while a higher PoC means reduced accessibility and suboptimal long-term behaviour due to clutter. 
In our experiments, PoC serves as an indicator of how effectively the agent’s actions improve long-term environment efficiency, capturing effects that standard task success or path-length metrics fail to account for.

\section{Long-term Efficiency Score}

\begin{figure*}[t]
\centering
\includegraphics[width=\linewidth]{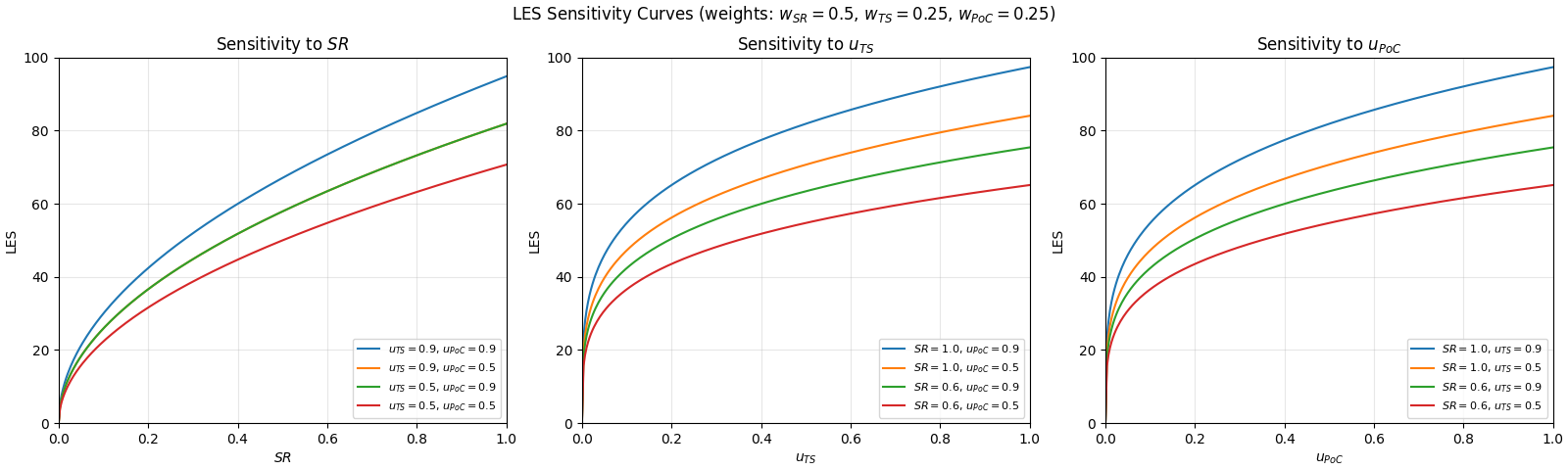}
\caption{LES sensitivity curves under the proposed multiplicative formulation ($w_{\text{SR}}=0.5$, $w_{\text{TS}}=0.25$, $w_{\text{PoC}}=0.25$). Each panel sweeps one factor (SR, $u_{\text{TS}}$, or $u_{\text{PoC}}$) while holding the others fixed at representative values; LES is plotted on the y-axis. The curves show that LES increases monotonically with improvements in any component, but is bounded when the remaining components are low, illustrating the intended trade-off: high scores require jointly high success, time efficiency, and long-term navigability.}\small
\label{F:les_sensitivity}
\end{figure*}

\label{sec:LES}
In lifelong interactive navigation, the agent must execute a sequence of navigation-manipulation tasks within the same evolving environment. 
Each object relocation persists and alters future navigability, meaning that decisions made for one task influence the difficulty, cost, and even feasibility of subsequent tasks. 
Existing metrics -- such as Success Rate (SR), Time Steps (TS), Path Length (PL), Success weighed by Path length (SPL), Success weighed by Time Steps (STS), and Price of Clutter -- capture only isolated aspects of performance. 
For example, SR does not reflect the cost of excessive detours or unnecessary object manipulation; TS alone cannot distinguish fast-but-myopic planners from methods that intentionally improve global structure; and PoC measures environment optimality but is insensitive to whether tasks are actually completed. 
Likewise, widely used embodied-AI metrics such as STS and SPL are designed for single-episode navigation with fixed environments -- they couple success to execution efficiency, but cannot capture how object relocation reshapes future traversability, nor do they penalize environment configurations that degrade long-horizon performance. 
Consequently, STS and SPL reward agents that reach the goal quickly in the current environment configuration, even if their manipulations make subsequent tasks substantially harder.
As a result, none of these metrics, taken independently, can evaluate long-horizon behavior or the quality of environment shaping. This motivates the need for a single, unified metric that integrates success, efficiency, and long-term environment optimality. \\

\noindent \textbf{Calculation:} Let $\mathrm{SR}$ denote the task success rate over an episode, $\mathrm{TS}$ the average number of time steps taken per episode, and $\mathrm{PoC}$ the Price of Clutter of the final modified environment. To produce a single higher-is-better score that is comparable across environment sizes and task horizons, we compute \emph{global} extrema over the full evaluation set (all methods, horizons, and floorplan groups) and convert $\mathrm{TS}$ and $\mathrm{PoC}$ into bounded utilities:
\[
u_{\mathrm{TS}} = \!1-\frac{\mathrm{TS}-\mathrm{TS}_{\min}}{\mathrm{TS}_{\max}-\mathrm{TS}_{\min}}, \qquad
u_{\mathrm{PoC}} = \!1-\frac{\mathrm{PoC}-\mathrm{PoC}_{\min}}{\mathrm{PoC}_{\max}-\mathrm{PoC}_{\min}},
\]
where $\mathrm{TS}_{\min},\mathrm{TS}_{\max},\mathrm{PoC}_{\min},\mathrm{PoC}_{\max}$ are global minima/maxima and the values of $u_{\text{TS}}$ and $u_{\text{PoC}}$ are normalized to $[0,1]$ for numerical robustness. We then define the \textbf{Long-term Efficiency Score (LES)} as a weighted geometric mean:
\[
\mathrm{LES} \;=\; 100 \cdot \mathrm{SR}^{w_{\mathrm{SR}}}\cdot (u_{\mathrm{TS}}+\epsilon)^{w_{\mathrm{TS}}}\cdot (u_{\mathrm{PoC}}+\epsilon)^{w_{\mathrm{PoC}}},
\qquad
w_{\mathrm{SR}}+w_{\mathrm{TS}}+w_{\mathrm{PoC}}=1,
\]
with a small $\epsilon$ (e.g., $10^{-8}$) for numerical stability. By construction, higher $\mathrm{SR}$ increases LES, while larger $\mathrm{TS}$ or $\mathrm{PoC}$ decrease LES via lower utilities. \\

\noindent \textbf{Rationale Behind the Formulation:} This exponentiated, multiplicative formulation is well-suited for lifelong interactive navigation:
\begin{itemize}
    \item \textbf{(1) Proper ``gating'' by success.} In lifelong settings, efficiency is only meaningful if tasks are actually completed. The multiplicative form ensures that low success cannot be compensated by favorable $\mathrm{TS}$ or $\mathrm{PoC}$, and $\mathrm{LES}=0$ when $\mathrm{SR}=0$.

    \item \textbf{(2) Balanced multi-objective trade-off.} Using a geometric mean (instead of an additive score) discourages methods that excel in only one dimension. Extremely poor performance in any one objective (e.g., highly cluttering the environment or taking very long trajectories) sharply reduces LES, reflecting the requirement that a lifelong agent must perform well \underline{simultaneously} on success, efficiency, and environmental optimality.

    \item \textbf{(3) Global normalization across scales.} Min--max normalization yields bounded utilities in $[0,1]$, making scores comparable across different environment sizes and obstacle densities, where raw $\mathrm{TS}$ and $\mathrm{PoC}$ can vary significantly in scale.

    \item \textbf{(4) Interpretable weighting via exponents.} The exponents $w_{\mathrm{SR}}, w_{\mathrm{TS}}$, and $w_{\mathrm{PoC}}$ directly control the relative importance of the three objectives while preserving scale invariance. In particular, increasing $w_{\mathrm{PoC}}$ places more emphasis on long-term navigability, whereas increasing $w_{\mathrm{TS}}$ emphasizes execution speed, and increasing $w_{\mathrm{SR}}$ prioritizes task completion.
\end{itemize}

\noindent \textbf{Interpretation:} High LES values indicate that an agent reliably completes the episode goals while remaining time-efficient and preserving global navigability of the environment (low $\mathrm{PoC}$). Moderate LES corresponds to agents that succeed but incur substantial detours or induce clutter configurations that degrade future navigation. Low LES values arise when the agent fails frequently, over-manipulates, or leaves the environment in a highly inefficient configuration. Because LES uses global normalization and bounded utilities, scores remain stable and comparable across floorplan groups and task horizons, providing a unified measure of long-term efficiency. \\

\noindent \textbf{Sensitivity Analysis:} The sensitivity curves in~\cref{F:les_sensitivity} illustrate how LES responds to controlled variations in each constituent factor—success rate (SR), time utility ($u_{\text{TS}}$), and clutter utility ($u_{\text{PoC}}$) -- while holding the remaining factors fixed at representative values (here $w_{\text{SR}}=0.5,\ w_{\text{TS}} = 0.25, w_{\text{{PoC}}}=0.25$). Each panel sweeps one variable from $0$ to $1$ and plots the resulting LES under multiple fixed settings of the other two variables, demonstrating that LES increases monotonically with improvements in any component and decreases when any component degrades.

A key takeaway is that LES enforces a meaningful trade-off rather than allowing any single metric to dominate. In the left panel, increasing SR steadily raises LES, but the achievable LES is capped when either $u_{\text{TS}}$ or $u_{\text{PoC}}$ is low, reflecting that high success alone is insufficient if the policy is slow or leaves the environment in a cluttered state. The middle and right panels show analogous behavior: improvements in efficiency $u_{\text{TS}}$ or environmental quality $u_{\text{PoC}}$ increase LES, but the gain is moderated when SR is low or the other utility is poor. Overall, the multiplicative (geometric-mean) form acts as a ``gating'' mechanism: strong performance requires simultaneously maintaining high task success, efficient execution, and low long-term clutter, aligning the metric with the goals of lifelong interactive navigation.

\section{Baseline Implementation Details}
\label{sec:baselines}
This section provides an extended description of the baselines used in our experiments. 
To ensure a fair comparison, \textbf{all baseline methods, except Ours (unk), operate under full environment observability}: each method is provided with the complete map of obstacles and objects at the start of the episode, including the positions of all task-relevant objects, all placement locations for each obstacle, and all obstacles that block navigation.
All baselines share the same embodiment and low-level actuation stack as our method: navigation follows a $0.25$\,m grid-based discretization available in ProcTHOR; manipulation is implemented via the standard \texttt{PickupObject} and \texttt{PutObject} primitives; and any failures at the level of gripper control, inverse kinematics, or grasp execution are handled by the low-level controller and does not terminate the episode. 
Thus, the baselines differ only in their high-level strategy for interacting with clutter and solving the navigation-manipulation objectives. \\

For clarity, we restate that the evaluation tasks are all object-to-object placement tasks (e.g., \texttt{Mug} $\rightarrow$ \texttt{Shelf}, \texttt{Vase} $\rightarrow$ \texttt{StudyTable}), and not navigate-to-region goals. 
All baselines attempt to complete the same $20$-task sequence.

\subsection{ADIN}
We implement ADIN~\citep{Wang2024AnIN} using the public codebase\footnote{\url{https://github.com/polkalian/ADIN/tree/main}} and retrain it end-to-end in our environment. 
Our goal is to preserve the original model and pipeline as faithfully as possible, introducing only the minimal changes required to match our environment setting and episode structure. \\

\noindent\textbf{Action primitives:} The original ADIN setup assumes an interaction primitive based on pushing. 
Since our setting requires object rearrangement via discrete pick-and-place operations, we replace the push primitive with a two-step manipulation sequence: \verb|PickupObject| followed by \verb|PutObject|. The policy continues to output actions over the environment’s discrete action set. \\

\noindent\textbf{Dynamic action/interface sizing:} The released implementation contains several hard-coded dimensions tied to a fixed action space (e.g., actor/critic output sizes of $10$ and $12$). 
We replace these constants with \verb|action_space.n| so the actor and critic heads automatically adapt to the action space in our environment. 
Similarly, we make the intent embedding size dynamic and replace the fixed uniform prior (originally $1/12$) with a uniform prior computed from the current number of actions. \\

\noindent\textbf{Reward adaptation:} We modify the reward function to better match long-horizon navigation under dense clutter and the two-stage nature of our tasks (retrieve $\rightarrow$ place). 
To prevent premature termination incentives in cluttered scenes, we remove the per-step penalty. 
The agent receives a positive geodesic reward as it moves (i) toward the goal object and then (ii) toward the target receptacle. We also reshape the sparse success reward $R_{\text{success}}$ to explicitly encourage sub-goal completion: the agent receives 
$R_{\text{success}}/2$ upon successfully picking up the correct object, and the remaining $R_{\text{success}}/2$ upon successfully placing it onto the target receptacle. \\

\noindent\textbf{Training protocol and data} We retrain ADIN end-to-end on our $10k$ training episodes from the cluttered multi-room dataset. 
Each task instance (a single object-to-receptacle command) constitutes one training episode. \\

\noindent\textbf{Observations and exploration:} Following the original assumption of privileged initialization in ADIN, the agent is provided full ground-truth environment information at episode start. 
As a result, it does not perform explicit exploration: all objects, obstacles, and occlusions are known at initialization. \\

\noindent We keep the rest of the method unchanged, including its mapping logic, pixel-level interaction predictor, frontier selection strategy, and short-range navigation module. Overall, this baseline constitutes a strong, faithful instantiation of the published ADIN approach, adapted only where necessary. 

\subsection{FastDownward}
We integrate a classical symbolic planner, Fast Downward (FD)~\citep{helmert2006fast}, into our AI2-THOR pipeline by automatically compiling each simulator state into a PDDL planning instance, solving it with FD, and translating the resulting plan back into executable AI2-THOR actions. \\

\noindent{\textbf{State abstraction:} At the beginning of each sub-task of an episode, we query the simulator for a discrete navigation graph using AI2-THOR’s \\ \verb|GetReachablePositions|. 
Each reachable pose is mapped to a symbolic location node, and directed adjacency relations are instantiated for the agent’s primitive motion model (we retain forward/back/left/right moves and rotations). 
We also extract all pickupable objects (including movable obstacles) and all candidate receptacles from the scene metadata and assign each a unique symbolic identifier. 
A task specification marks (i) the set of goal objects, and (ii) the set of valid target receptacles.
We re-query the simulator after each successful task for writing the new PDDL state. \\

\noindent\textbf{PDDL domain modeling:} We define a STRIPS-style PDDL domain capturing navigation, grasping, and placement under AI2-THOR’s constraints. 
The symbolic state includes the agent location and orientation, whether the hand is empty, whether an object is held, and object placement on receptacles. 
We explicitly disallow placing objects onto the floor; pickup from the floor is modeled by requiring the agent to be adjacent to the object’s location. 
To ensure compatibility with the Fast Downward configuration used in our pipeline, we avoid conditional effects by enumerating orientation-specific rotation operators (e.g., rotateleft-north, rotateleft-west, $\cdots$) instead of a single action with conditional effects. \\

\noindent\textbf{PDDL problem generation:} In each episode, we generate a PDDL problem file for the current task that instantiates:
\begin{itemize}
    \item the navigation graph over reachable locations,

    \item the initial agent pose (at-agent, facing),

    \item the initial placements of objects,

    \item the task-specific predicates identifying goal objects and target receptacles,

    \item the goal condition, which requires that any one goal object is placed on any one target receptacle (encoded as an existential-over-candidates using a compiled predicate such as delivered).
\end{itemize}

\noindent Alongside each PDDL problem, we write a symbol-to-simulator mapping file (\verb|.symmap.json|) that records the correspondence between PDDL symbols (objects/receptacles) and AI2-THOR objectIds. 
This mapping is later used to ground planned actions in the simulator.
After each successful task, the PDDL problem file is re-generated for the new environment state. \\

\noindent\textbf{Planning:} We invoke Fast Downward on each generated domain/problem pair using a standard heuristic search configuration (e.g., \verb|astar(lmcut())|). 
We configure FD to write the resulting plan to a deterministic plan file (the SAS plan format). 
We implement a time limit of $1800.0$ seconds for the FD planner.
If FD fails to find a plan, we treat the episode as unsolved by the planning baseline. \\

\noindent\textbf{Plan execution in AI2-THOR:} We parse the SAS plan into a sequence of symbolic operator calls and translate each to an AI2-THOR action. 
Navigation actions map directly to THOR primitives (\verb|MoveAhead|, \verb|MoveBack|, \verb|MoveLeft|, \verb|MoveRight|, \verb|RotateLeft|, \verb|RotateRight|). 
Manipulation actions are grounded using the stored symbol map: \verb|PickupObject*| is executed with the corresponding \verb|objectId|, and \verb|PutObject| is executed with \verb|receptacleObjectId|. 
Because rotation operators are orientation-specific in PDDL, all of them map to the same THOR rotation primitive at execution time. 
The resulting executable plan is saved as a JSON list of THOR actions, which can be replayed in the simulator. \\

\noindent\textbf{Outcome:} This integration provides a fully classical planning baseline that uses the simulator only to build a discrete abstraction and then plans in symbolic space, enabling controlled comparisons against learning-based methods while using the same low-level action interface.

\subsection{InterNav}
\textbf{Original capability:} InterNav~\citep{zeng2021pushing} (officially released as \verb|ObjPlace|) is designed for single-room interactive navigation using pushing actions on small tabletop or floor obstacles. The original action space consists of:
\begin{itemize}
    \item forward/turn navigation actions,
    \item \emph{push} actions, applied at pixel-level predicted contact points.
\end{itemize}

\noindent\textbf{Action-space adaptation:} Since our problem setting requires object rearrangement through pick-place actions rather than pushing, we replace the push primitive with \texttt{PickupObject} + \texttt{PutObject}. No modifications are made to the architecture. \\

\noindent\textbf{Reward structure adaptation:} We adapted the original reward structure to better suit our specific task requirements. To prevent the agent from prematurely terminating episodes while navigating dense clutter, we eliminated the step penalty. The task reward is distributed across the two distinct phases of the episode - navigation phase and manipulation phase. During navigation, when the agent moves towards the goal object and subsequently towards the target receptacle, the agent receives a positive geodesic reward identical to the original InterNav formulation. The sparse success reward ($R_{\text{success}}$) is split to encourage sub-goal completion: the agent receives $R_{\text{success}}/2$ upon successfully picking up the correct object, and the remaining $R_{\text{success}}/2$ upon placing the object on the target receptacle. \\

\noindent\textbf{Training setup:} InterNav is re-trained end-to-end on our $10$k training episo-\\des of cluttered multi-room dataset. 
Each task instance (one object-to-receptacle command) serves as a single training episode.
Training environments are sampled with obstacle density uniformly drawn from $d=0$ to the maximum observed in our test split, ensuring a matched distribution. \\

\noindent\textbf{Environment knowledge:} InterNav receives full ground-truth environment information. 
It does not require exploration since objects, obstacles, and all occlusions are known at initialization. \\

\noindent\textbf{Software implementation:} Aside from the change in action primitives (push $\rightarrow$ pick-place), we do not modify: its mapping logic, its pixel-level interaction predictor, its frontier selection, or its short-range navigation module.\\

\noindent Thus the baseline represents a faithful, strong instantiation of the published method, adapted only minimally to match the manipulation capabilities of our setting.

\subsection{Always Detour}
This baseline represents an extreme non-interactive strategy. \\

\noindent\textbf{Interaction rule:} The agent is forbidden from manipulating obstacles. 
If an obstacle lies on the current shortest path to the target object, the agent immediately declares failure. \\

\noindent\textbf{Path computation:} Shortest paths are computed using \verb|networkx|’s Dijkstra solver on the full ground-truth graph. Because all floorplan geometry and all obstacles are known, the path planner never needs to explore. \\

\noindent\textbf{Consequences:}
\begin{itemize}
    \item Success depends purely on whether an entirely obstacle-free path exists.
    \item Never interacts with obstacles.
    \item Adversarially placed mild clutter can cause early termination.
    \item TS can be low (episodes often end quickly) but the low LES score reveals that this is purely due to avoiding meaningful interaction.
\end{itemize}

\subsection{Always Interact}
This baseline removes \emph{every} encountered obstacle that blocks the shortest path to the current goal. \\

\noindent\textbf{Interaction rule:} Upon encountering an obstacle, the agent:
\begin{enumerate}
    \item picks up the obstacle using \texttt{PickupObject},
    \item navigates to and places it at the nearest available location that does \emph{not} block any path in the free-space graph (computed using ground-truth map),
    \item recomputes the shortest path and continues.
\end{enumerate}

\noindent\textbf{Ground-truth reasoning:} The baseline has complete access to all traversability information and all valid receptacle surfaces. 
Thus it can deterministically choose a placement that is globally non-blocking. \\

\noindent\textbf{Characteristics:}
\begin{itemize}
    \item Removes all obstacles on the path, regardless of the consideration for long-term tasks.
    \item PoC is near-ideal because the agent aggressively removes everything on its immediate route.
    \item TS is high due to excessive manipulation.
\end{itemize}

\subsection{Clean + Shortest Path (Clean + S/P)}
This is an extreme over-manipulation baseline. \\

\noindent\textbf{Interaction rule:} Before attempting any task, the agent identifies \emph{all} movable obstacles in the environment -- whether or not they lie on the expected path -- and removes them one by one. \\

\noindent\textbf{Placement rule:} Objects are placed at the nearest available location that preserves global traversability, identical to Always Interact’s placement criterion. \\

\noindent\textbf{Execution:} After full cleanup:
\begin{itemize}
    \item all obstacles have been removed from free space,
    \item the environment is in a globally optimal state ($\mathrm{PoC}\approx 1.00$),
    \item the agent executes all remaining tasks using shortest paths.
\end{itemize}

\noindent\textbf{Characteristics:}
\begin{itemize}
    \item TS is extremely high due to large-scale cleanup.
    \item Over-interacts with obstacles in the environment.
    \item Environment optimality is maximal, but reasoning is naive.
\end{itemize}

\subsection{Ours (Known Environment)}
This variant isolates the reasoning component of our framework. \\

\noindent\textbf{Access to full state:} The LLM-based planner receives the complete scene graph at the start: all obstacles, all objects, all receptacles, and all traversability. \\

\noindent\textbf{No exploration:} The agent does not need to look for objects or reconstruct the map. 
The policy focuses solely on solving the sequence of tasks while strategically removing obstacles only when they improve long-term connectivity. \\

\noindent\textbf{Interpretation:} This is an upper bound on our method’s structural reasoning ability, free from challenges of partial observability.

\subsection{Ours (Unknown Environment)}
This is the full framework introduced in the main paper. \\

\noindent\textbf{Sensing:} The agent perceives the world exclusively through its front-facing cameras. Unlike all other baselines:
\begin{itemize}
    \item it begins with no map,
    \item does not know obstacle positions,
    \item does not know object locations,
    \item must discover all task-relevant elements actively.
\end{itemize}

\noindent\textbf{Active perception + LLM reasoning:} The LLM receives the incrementally built scene graph and jointly chooses:
\begin{itemize}
    \item where to explore next,
    \item which objects to move,
    \item where to relocate them,
    \item how to sequence navigation, manipulation, and exploration.
\end{itemize}

\noindent\textbf{Low-level control:} All navigation and manipulation actions are executed using the same primitives as in simulation, so comparison with baselines is fair. \\

\noindent\textbf{Episode termination:} The episode ends when:
\begin{itemize}
    \item all $20$ object-placement tasks are completed, or
    \item the global \texttt{max\_steps} threshold is exceeded.
\end{itemize}

\noindent This baseline is the only one that must solve the \emph{lifelong, partially observed, cluttered} problem as formulated in our paper.

\section{Baseline Comparison (Extended Results)}
\label{sec:base_comp_ext}
In this section, we report two additional evaluations that highlight qualitative differences in how each baseline behaves over long-horizon episodes:
\begin{itemize}
    \item the final Interaction Efficiency (IE) for each baseline across binned floorplans, and

    \item the Lifelong Environment Score (LES) plotted as a function of floorplan size.
\end{itemize}
These two metrics together quantify how selectively a method manipulates obstacles (IE) and how effectively its decisions maintain long-term environment optimality (LES).

\subsection{LES Across Increasing Environment Sizes}

\begin{figure*}[t]
\centering
\includegraphics[width=\linewidth]{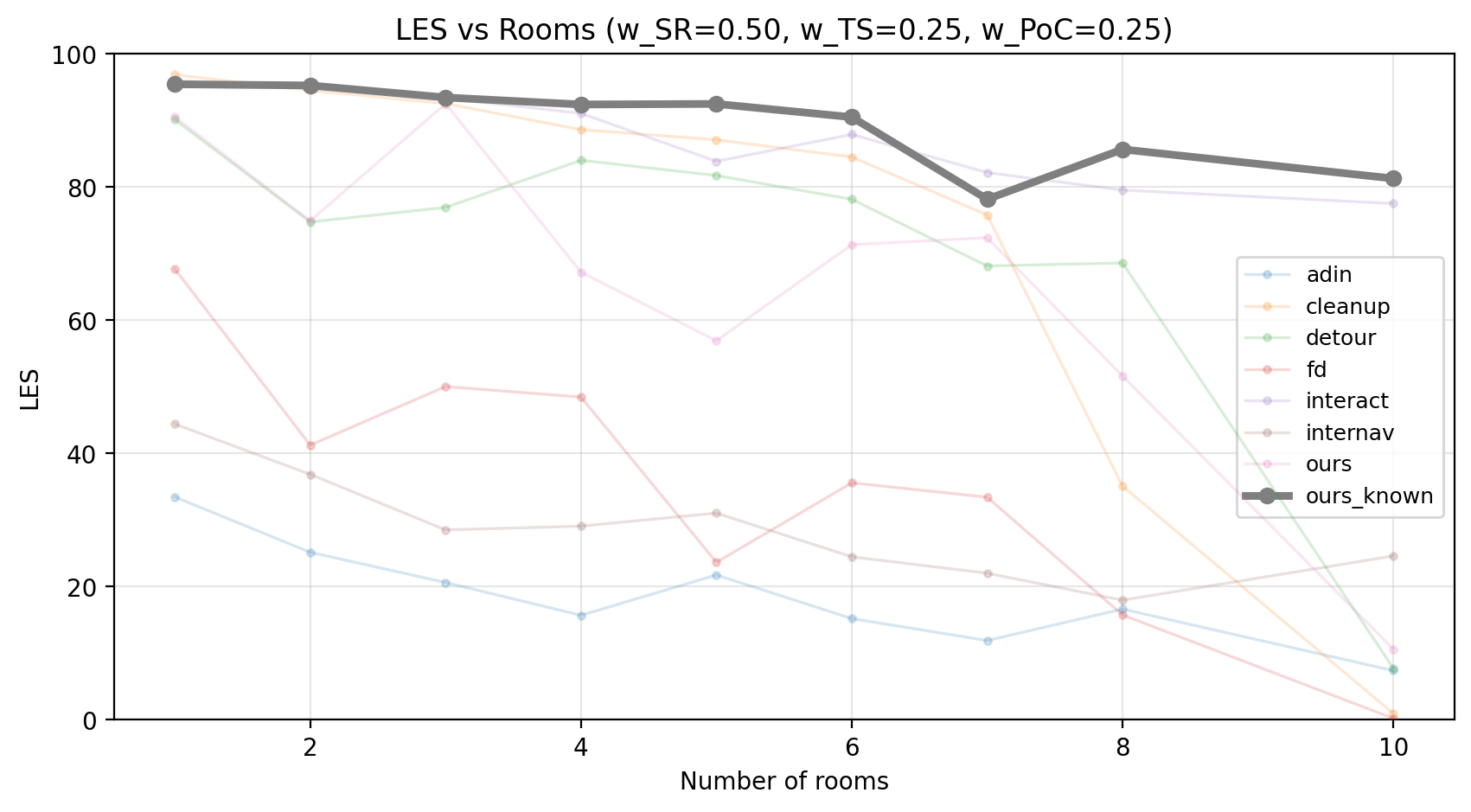}
\caption{Lifetime Efficiency Score (LES) plot for floorplans ranging from $1$ to $10$ rooms. Higher is better. In small environments, nearly all baselines appear competitive because suboptimal actions -- such as excessive detours (Detour) or indiscriminate manipulation (Interact, Cleanup) -- incur limited long-term cost. As spatial complexity increases, these naïve policies collapse: detour-heavy strategies suffer from compounding path inefficiency, while cleanup-heavy strategies incur large manipulation overhead. InterNav and ADIN fail to scale due to brittle obstacle-handling assumptions. Our method remains consistently stable across environment sizes. The widening performance gap highlights the need for selective, globally informed interaction when navigating lifelong, cluttered environments.}\small
\label{F:bext_les}
\end{figure*}

To understand how each method scales with environment complexity, \cref{F:bext_les} plots the Lifelong Environment Score (LES) for all baselines across floorplans containing $1$ to $10$ rooms. 
LES jointly captures (i) task success across the episode, (ii) the long-term navigability of the environment as measured by Price of Clutter (PoC), and (iii) the executed time steps (TS). 
Thus, higher LES indicates methods that not only solve tasks but also preserve global structure and efficiency across the entire task sequence. \\

\noindent\textbf{Small environments (1–3 rooms):} In compact spaces, nearly all baselines exhibit moderate-to-high LES. 
This occurs because the penalty for suboptimal actions is inherently low: even aggressive strategies such as Always Interact or Clean+S/P can over-manipulate the scene with little long-term cost, while purely reactive strategies such as Always Detour can avoid manipulation entirely without incurring long detours. 
The limited spatial extent caps PoC, shortens detours, and reduces compounding errors, making the differences between methods less pronounced. 
Nevertheless, our method and the oracle variant (Ours-known) already show strong LES due to more selective manipulation and shorter time-to-completion. \\

\noindent\textbf{Medium environments (4–6 rooms):} As spatial layout grows, naïve strategies begin to diverge. 
Always Detour and InterNav collapse quickly because never manipulating obstacles leads to increasingly long detours and frequent infeasible plans, driving LES downward. 
Always Interact and Clean+S/P degrade as well: although they can maintain low PoC by removing many obstacles, the large TS cost of exhaustive cleanup drags LES down sharply. 
In contrast, our method maintains high LES by performing targeted, high-impact rearrangements while avoiding both excessive cleanup and unbounded detours. 
This selective behavior retains good PoC and strong SR with considerably lower TS than cleanup-based baselines. \\

\noindent\textbf{Large environments (7–10 rooms):} In large-scale floorplans, the advantages of long-term reasoning become essential. 
All non-reasoning baselines experience severe LES deterioration. 
Detour and InterNav fail almost entirely due to compounding detours, disconnected regions, and task infeasibility. 
Always Interact and Clean+S/P suffer from the opposite failure mode: the cost of clearing many distant or irrelevant obstacles explodes as room count increases, yielding high SR but extremely poor TS and diminished LES.  \\

\noindent Overall, LES reveals a crucial scaling trend:
In small environments, many baselines appear competitive because the cost of poor structural decisions is inherently limited.
As environments grow, only methods that explicitly reason about long-term connectivity, clutter placement, and selective manipulation maintain strong performance.
Our approach is the only non-oracle method whose LES curve remains significantly high and stable across all room counts, demonstrating effective lifelong planning and environment shaping.

\begin{table}[t]
\centering
\small
\setlength{\tabcolsep}{4pt}
\renewcommand{\arraystretch}{1.05}

\begin{tabular}{l|c|c|c}
\toprule
\textbf{\% Interactions} & \textbf{1--3 rooms} & \textbf{4--6 rooms} & \textbf{7--10 rooms} \\
\midrule
Ours (known)     & 28.67 & 37.33 & 31.67 \\
Ours (unk)       & 36.61 & 38.48 & 32.09 \\
Always Interact  & 100.00& 100.00& 100.00 \\
Always Detour    & 0.00  & 0.00  & 0.00 \\
Clean + S/P      & 158.00& 161.61& 182.47 \\
\bottomrule
\end{tabular}

\caption{Percentage of interactions (IE) executed by each baseline across the three floorplan bins.}\small
\label{tab:bext_ie}
\end{table}

\noindent To complement the LES analysis, we examine the percentage of interactions (IE) executed by each baseline across the three floorplan bins (Table~\ref{tab:bext_ie}). IE measures how often an agent chooses to manipulate obstacles relative to the number of obstacles it encounters. Since excessive manipulation is costly in both time and long-term structural distortion, IE captures a core determinant of high LES. \\

\noindent A clear trend emerges: methods with extreme interaction policies achieve the lowest LES in large environments. 
\textbf{Always Interact} performs $100\%$ manipulation regardless of context; this directly explains its sharp LES drop as room count increases (Fig.~\ref{F:bext_les}). 
\textbf{Clean + S/P} is extremely aggressive in obstacle manipulation -- it removes all obstacles before solving any of the tasks -- leading to IE values greater than $150\%$ as it manipulates more obstacles than it ever would encounter along optimal routes. 
These policies appear competitive in small environments, where their heavy-handed behavior does not incur substantial penalties; however, as free-space geometry becomes more complex, such over-interaction drastically harm long-horizon efficiency, reflected in steep LES decay. \\

\noindent Conversely, \textbf{Always Detour} maintains $0\%$ IE by design. 
While this avoids direct manipulation costs, it forces the agent into long navigation routes whenever clutter blocks high-connectivity regions. 
In larger floorplans, these detours accumulate geometric inefficiencies that outweigh any benefits of interaction frugality, again explaining its lowering LES values as the number of rooms increases. \\

\noindent Our methods -- \textbf{Ours (unk)} and \textbf{Ours (known)} -- have IE values in moderate range ($28$--$38\%$), increasing slightly with floorplan size as more rooms imply a higher chance of encountering globally central obstacles. 
Unlike the over-reactive baselines, our approach interacts only when doing so meaningfully improves long-term navigability -- the behavior rewarded by LES. 
The LES curves reflect this: while all baselines degrade as room count increases, our approach maintains significantly higher efficiency, with \textbf{Ours (known)} forming an approximate upper bound on selective, globally informed rearrangement. \\

\noindent Together, these results show that only planners capable of evaluating global structure -- rather than reacting myopically to local blockages -- achieve this balance consistently across floorplan scales.

\begin{table*}[t]
\centering
\resizebox{\textwidth}{!}{%
\begin{tabular}{l|c c c c|c c c c|c c c c}
\toprule

\textbf{Obstacle Density} & 
\multicolumn{4}{c|}{\textbf{$\mathbf{1-3}$ rooms}} & 
\multicolumn{4}{c|}{$\mathbf{4-6}$\textbf{ rooms}} & 
\multicolumn{4}{c}{$\mathbf{7-10}$\textbf{ rooms}} \\
\cmidrule(lr){2-5} \cmidrule(lr){6-9} \cmidrule(lr){10-13}
 & SR & PoC & TS & LES
 & SR & PoC & TS & LES
 & SR & PoC & TS & LES\\

\midrule
$\mathbf{d=0.5}$ \\
CoReLIN (known) & $95.61$ & $1.10$ & $931.45$ & $94.95$ & $100.00$ & $1.11$ & $1769.19$ & $\mathbf{92.27}$ & $94.85$ & $1.30$ & $2384.24$ & $\mathbf{84.47}$\\

CoReLIN (unk) & $86.52$ & $1.31$ & $1678.06$ & $84.80$ & $79.50$ & $1.81$ & $3512.33$ & $65.56$ & $69.10$ & $2.79$ & $4419.97$ & $45.15$\\

ADIN & $15.50$ & $1.48$ & $809.53$ & $37.09$ & $5.90$ & $2.11$ & $1879.02$ & $19.91$ & $3.1$ & $1.73$ & $3025.72$ & $13.76$\\ 

Always Interact & $100.00$ & $1.07$ & $1026.18$ & $\mathbf{96.85}$ & $100.00$ & $1.29$ & $2032.05$ & $89.10$ & $100.00$ & $1.35$ & $3359.55$ & $\underline{78.66}$\\

Always Detour & $87.88$ & $1.37$ & $707.06$ & $89.89$ & $86.97$ & $1.68$ & $1240.48$ & $84.10$ & $73.48$ & $2.65$ & $1620.52$ & $65.61$\\

Clean + S/P & $100.00$ & $1.00$ & $1107.71$ & $\underline{96.68}$ & $100.00$ & $1.00$ & $2126.58$ & $\underline{90.74}$ & $100.00$ & $1.00$ & $4320.03$ & $69.92$\\

\midrule 
$\mathbf{d=1.0}$ \\
CoReLIN (known) & $94.55$ & $1.50$ & $1044.55$ & $89.20$ & $97.73$ & $1.25$ & $1911.97$ & $\mathbf{88.37}$ & $100.00$ & $1.23$ & $2435.48$ & $\mathbf{86.56}$\\ 

CoReLIN (unk) & $88.94$ & $1.48$ & $1339.36$ & $83.87$ & $81.67$ & $1.81$ & $3211.20$ & $67.25$ & $62.27$ & $2.79$ & $4497.80$ & $49.99$\\ 

ADIN & $13.81$ & $1.81$ & $602.86$ & $37.42$ & $4.76$ & $1.43$ & $1453.61$ & $20.03$ & $2.86$ & $2.90$ & $2178.04$ & $12.21$\\

Always Interact & $100.00$ & $1.14$ & $1108.88$ & $\underline{94.18}$ & $100.0$ & $1.13$ & $2416.67$ & $\underline{87.19}$ & $100.00$ & $1.36$ & $3974.88$ & $\underline{74.73}$\\ 

Always Detour & $90.45$ & $1.96$ & $665.64$ & $85.12$ & $83.18$ & $2.15$ & $1115.27$ & $78.26$ & $64.09$ & $3.43$ & $1381.61$ & $54.99$\\ 

Clean + S/P & $100.00$ & $1.00$ & $1200.48$ & $\mathbf{94.38}$ & $100.00$ & $1.00$ & $2622.94$ & $86.65$ & $100.00$ & $1.00$ & $5169.58$ & $61.86$\\ 

\midrule

$\mathbf{d=2.0}$ \\
CoReLIN (known) & $94.55$ & $2.56$ & $915.76$ & $90.59$ & $92.88$ & $2.46$ & $2014.21$ & $\mathbf{83.54}$ & $92.88$ & $2.49$ & $2904.09$ & $\mathbf{77.67}$\\

CoReLIN (unk) & $81.88$ & $3.19$ & $1592.34$ & $79.23$ & $55.15$ & $3.67$ & $3135.45$ & $56.72$ & $48.79$ & $4.55$ & $4773.21$ & $42.46$\\

ADIN & $15.6$ & $3.77$ & $774.95$ & $35.88$ & $5.1$ & $4.95$ & $1953.90$ & $18.00$ & $3.7$ & $5.66$ & $2982.57$ & $13.35$\\ 

Always Interact & $100.00$ & $1.41$ & $1314.16$ & $\mathbf{92.97}$ & $100.00$ & $1.72$ & $2953.97$ & $\underline{81.35}$ & $100.00$ & $1.45$ & $5135.61$ & $\underline{62.58}$\\

Always Detour & $84.85$ & $3.30$ & $386.48$ & $85.92$ & $41.67$ & $4.66$ & $541.42$ & $57.34$ & $38.18$ & $5.40$ & $714.82$ & $52.00$\\

Clean + S/P & $100.00$ & $1.00$ & $1457.42$ & $\underline{91.89}$ & $100.00$ & $1.00$ & $3609.36$ & $77.40$ & $100.00$ & $1.00$ & $6879.50$ & $35.54$\\

\bottomrule
\end{tabular}
}
\caption{Comparison of our approach with baselines across different floorplans under varying obstacle densities.}
\label{tab:abl_obs_dense}
\end{table*}

\section{Obstacle Density Experiments: Comparison with Baselines}
\label{sec:dense_perf}

To evaluate robustness under different levels of clutter, we compare all methods across three obstacle densities: $d=0.5$ (sparser than default), $d=1.0$ (default density), and $d=2.0$ (highly cluttered scenes), while keeping the environments and task sequences fixed. Across all settings, we report Success Rate (SR), Path over Cost (PoC), Time Steps (TS), and our unified long-horizon metric LES. The results are shown in \cref{tab:abl_obs_dense}. \\

\noindent A consistent trend across all methods is that increasing obstacle density makes the task substantially harder: TS and PoC generally increase, while LES degrades, especially in medium and large floorplans where clutter compounds long-horizon navigation and manipulation costs. This degradation is particularly severe for heuristic baselines such as ADIN and Always Detour, whose LES drops sharply as scenes become denser and larger. Always Interact remains competitive in the easiest settings due to its high SR, but incurs increasingly large manipulation overhead as density grows. In contrast, CoReLIN maintains strong performance across densities, showing that selectively removing structurally important obstacles is more effective than either indiscriminate interaction or purely reactive detouring. \\

\noindent In sparse environments ($d=0.5$), the task is naturally easier for all methods, and the gap between methods is smallest in the $1-3$ room regime. In these easy layouts, short travel distances reduce the benefit of long-horizon reasoning, so simple strategies such as always interacting or aggressively clearing obstacles can remain competitive. However, as the number of rooms increases, CoReLIN becomes the strongest method. Even at low obstacle density, larger floorplans still require reasoning about which obstacles will create downstream bottlenecks, and our selective interaction policy is better able to account for these long-term effects than reactive detouring or indiscriminate cleanup. This shows that the benefit of our approach is not tied only to extreme clutter, but already emerges once tasks require nontrivial long-horizon planning. \\

\noindent At the default density ($d=1.0$), the same trend becomes more pronounced. In small floorplans, CoReLIN remains highly competitive, though the margin over simpler baselines is limited because these environments are still relatively easy and leave little room for poor long-term decisions to accumulate. In contrast, in medium and large floorplans, CoReLIN clearly performs best overall. As clutter and task horizon increase together, heuristic methods begin to suffer from their fixed behavior: always interacting incurs unnecessary manipulation cost, while detour-based strategies become inefficient once navigation options narrow. CoReLIN, by selectively removing only task-critical obstacles, scales much more gracefully and achieves the strongest overall balance between success, navigation efficiency, and interaction cost. \\

\noindent In highly cluttered scenes ($d=2.0$), the advantage of our approach becomes most evident. All baselines degrade as dense clutter increases traversal cost, restricts free-space connectivity, and raises the cost of unnecessary manipulation. Yet CoReLIN remains the best-performing practical approach, especially in medium and hard floorplans where long-horizon dependencies are most severe. Reactive or exhaustive strategies become increasingly inefficient in this regime: detouring is less viable when alternative paths are limited, while blanket cleanup overpays in manipulation cost. In contrast, CoReLIN continues to focus interaction on the small set of obstacles that matter most for future task progress, allowing it to retain the strongest performance as environments become both larger and more cluttered. \\

\noindent The unknown-map variant follows the same overall trend but performs below the known-map version due to the added burden of exploration under partial observability. Its degradation is most visible in medium and large scenes, where dense clutter both occludes task-relevant objects and restricts mobility. Nevertheless, it remains substantially stronger than weaker baselines such as ADIN across all settings, confirming that the underlying reasoning policy remains effective even when map information is unavailable. \\

\noindent Overall, these experiments reinforce the main conclusion of our work: although all methods degrade as obstacle density increases, CoReLIN is the most robust practical approach under increasing clutter. It is only narrowly outperformed in the easiest small-floorplan settings, where path lengths are too short for long-horizon penalties to dominate, but it consistently achieves the best performance in medium and hard floorplans across all densities.

\section{Clutter Generation Policy}
\label{sec:clutter_gen_policy}

In this section, we evaluate the sensitivity of our results to the clutter generation policy used to populate the environment with movable obstacles. 
Specifically, under the fixed episode-horizon setting $g=20$, we focus on the strongest baselines reported in the main paper -- Always Detour, Always Interact, and Clean + S/P—and omit ADIN, InterNav, and FastDownward since they were substantially weaker under the same $g=20$ evaluation setting, and do not affect the conclusions of this analysis.
To ensure a controlled comparison, all methods are provided ground-truth knowledge of the environment and are evaluated under identical environment and task configurations, differing only in how clutter is instantiated. 
Across all considered policies, our approach consistently outperforms the baselines, indicating that the gains are not tied to a particular clutter distribution but instead reflect robust performance under diverse environment generation processes.

\begin{table*}[t]
\centering
\resizebox{\textwidth}{!}{%
\begin{tabular}{l|c c c c|c c c c|c c c c}
\toprule

\textbf{Clutter Generation} & 
\multicolumn{4}{c|}{\textbf{$\mathbf{1-3}$ rooms}} & 
\multicolumn{4}{c|}{$\mathbf{4-6}$\textbf{ rooms}} & 
\multicolumn{4}{c}{$\mathbf{7-10}$\textbf{ rooms}} \\
\cmidrule(lr){2-5} \cmidrule(lr){6-9} \cmidrule(lr){10-13}
 \textbf{Policy} & SR & PoC & TS & LES$\uparrow$
 & SR & PoC & TS & LES$\uparrow$
 & SR & PoC & TS & LES$\uparrow$\\

\midrule 
\textbf{Clearance} \\
CoReLIN (known) & $98.80$ & $1.30$ & $943.61$ & $96.12$ & $100.00$ & $1.28$ & $1656.70$ & $\mathbf{95.45}$ & $100.00$ & $1.28$ & $2135.33$ & $\mathbf{94.09}$\\ 

Always Interact & $100.0$ & $1.12$ & $973.45$ & $\mathbf{98.38}$ & $97.70$ & $1.09$ & $1775.94$ & $\underline{94.70}$ & $100.00$ & $1.14$ & $2821.27$ & $92.61$\\ 

Always Detour & $96.50$ & $1.28$ & $774.85$ & $94.92$ & $90.90$ & $1.83$ & $1261.27$ & $84.39$ & $100.00$ & $1.23$ & $1989.33$ & $\underline{93.29}$\\ 

Clean + S/P & $100.0$ & $1.00$ & $1036.76$ & $\underline{97.20}$ & $100.00$ & $1.00$ & $2036.48$ & $94.32$ & $97.90$ & $1.00$ & $4162.94$ & $84.03$\\ 

\midrule 
\textbf{Degree Centrality} \\
CoReLIN (known) & $100.00$ & $1.25$ & $957.39$ & $97.26$ & $100.00$ & $1.22$ & $1584.24$ & $\mathbf{94.93}$ & $99.7$ & $1.46$ & $2152.94$ & $\mathbf{90.08}$\\ 

Always Interact & $100.00$ & $1.09$ & $1071.97$ & $\mathbf{97.90}$ & $99.2$ & $1.07$ & $1850.82$ & $\underline{94.02}$ & $100.00$ & $1.61$ & $2851.76$ & $84.42$\\ 

Always Detour & $97.00$ & $1.41$ & $834.09$ & $93.99$ & $96.5$ & $1.14$ & $1319.79$ & $93.87$ & $94.10$ & $1.13$ & $1916.39$ & $\underline{89.46}$\\ 

Clean + S/P & $100.00$ & $1.00$ & $1160.30$ & $\underline{97.85}$ & $96.2$ & $1.00$ & $2064.73$ & $90.34$ & $100.00$ & $1.00$ & $4066.42$ & $77.05$\\ 

\midrule
\textbf{Load Centrality} \\
CoReLIN (known) & $100.00$ & $1.35$ & $986.55$ & $96.84$ & $100.00$ & $1.17$ & $1764.45$ & $\mathbf{93.29}$ & $98.00$ & $2.34$ & $1978.23$ & $\mathbf{86.15}$\\ 

Always Interact & $100.00$ & $1.08$ & $1110.00$ & $\underline{97.09}$ & $100.00$ & $1.07$ & $1967.58$ & $92.37$ & $99.50$ & $1.07$ & $3148.17$ & $\underline{83.26}$\\ 

Always Detour & $91.50$ & $1.73$ & $693.15$ & $91.01$ & $80.30$ & $2.18$ & $1134.85$ & $79.68$ & $85.90$ & $1.58$ & $1924.41$ & $81.83$\\ 

Clean + S/P & $100.00$ & $1.00$ & $1061.58$ & $\mathbf{97.49}$ & $100.00$ & $1.00$ & $1946.82$ & $\underline{92.59}$ & $100.00$ & $1.00$ & $4210.30$ & $69.50$\\ 

\midrule 
\textbf{Betweenness Centrality} \\
CoReLIN (known) & $94.55$ & $1.50$ & $1044.55$ & $89.20$ & $97.73$ & $1.25$ & $1911.97$ & $\mathbf{88.37}$ & $100.00$ & $1.23$ & $2435.48$ & $\mathbf{86.56}$\\ 

Always Interact & $100.00$ & $1.14$ & $1108.88$ & $\underline{94.18}$ & $100.0$ & $1.13$ & $2416.67$ & $\underline{87.19}$ & $100.00$ & $1.36$ & $3974.88$ & $\underline{74.73}$\\ 

Always Detour & $90.45$ & $1.96$ & $665.64$ & $85.12$ & $83.18$ & $2.15$ & $1115.27$ & $78.26$ & $64.09$ & $3.43$ & $1381.61$ & $54.99$\\ 

Clean + S/P & $100.00$ & $1.00$ & $1200.48$ & $\mathbf{94.38}$ & $100.00$ & $1.00$ & $2622.94$ & $86.65$ & $100.00$ & $1.00$ & $5169.58$ & $61.86$\\ 

\bottomrule
\end{tabular}
}
\caption{Comparison of our approach with baselines across different floorplans across varying clutter generation policies.}
\label{tab:abl_cl_gen}
\end{table*}

\subsection{Clearnace-based}
Under the clearance-based clutter generation policy, we first compute a clearance value at each reachable coordinate/node, defined as the (normalized) distance to its nearest obstacle in the underlying environment. 
Concretely, we measure the distance from each node to the closest obstacle, normalize clearance to $[0,1]$ within each environment, and then occlude a fixed percentage of nodes by instantiating clutter at sampled coordinates (following the same occlusion rate protocol as in the original dataset). 
This policy differs from our standard clutter generation in the main paper, which is betweenness-centrality–based: rather than targeting graph-theoretic ``connectors'' that lie on many shortest paths and thus creating bottlenecks, clearance-based clutter targets regions based on geometric proximity to obstacles (i.e., how ``open'' or ``tight'' local free space is), producing a qualitatively different distribution of occlusions. \\

\noindent As indicated in~\cref{tab:abl_cl_gen}, cross the environment sizes, the results under clearance-based clutter reinforce the same conclusion as the betweenness-centrality setting: CoReLIN (known) consistently matches or exceeds the strongest baselines. 
In particular, CoReLIN achieves near-ceiling success rate throughout and strong LES, while baselines that either always interact or always detour do not consistently improve—and can lag in efficiency metrics as environments scale. 
Overall, these results indicate that the performance gains of our approach are not specific to a centrality-targeted clutter distribution, but persist under a geometry-driven occlusion process that changes where clutter appears and how it impacts navigation and interaction.

\subsection{Degree centrality based}
Under the degree-centrality–based clutter policy, we represent the navigable free space as a graph (nodes as reachable coordinates, edges as feasible local transitions) and compute each node’s degree centrality (i.e., the number of adjacent navigable neighbors, normalized by the maximum possible degree). 
We then instantiate clutter by occluding a fixed percentage of nodes sampled according to this degree-centrality score (biasing occlusions toward locally well-connected ``junction-like'' regions). 
This differs from our standard clutter generation in the main paper, which is betweenness-centrality–based: betweenness targets nodes that lie on many global shortest paths (structural bottlenecks/connectors), whereas degree centrality is a local measure of connectivity that emphasizes intersections and locally expansive regions rather than global bridge points. \\

\noindent As shown in \cref{tab:abl_cl_gen}, across all environment sizes ($1-3$, $4-6$, and $7-10$ rooms), the results follow the same qualitative trend as the betweenness-centrality setting: CoReLIN (known) remains the strongest overall method under degree-centrality–based clutter. 
In particular, CoReLIN achieves near-ceiling success rates ($\approx100\%$) while maintaining higher efficiency (LES) as environments scale (e.g., best LES in $4-6$ rand $7-10$ rooms), whereas the best heuristic baselines (Always Interact / Always Detour / Clean + S/P) degrade more noticeably in larger environments. 
Overall, this confirms that our gains are not tied to the specific betweenness-based clutter distribution used in the main paper, but persist under a distinct, locally connectivity-driven clutter policy.

\subsection{Load centrality based}
Under the load-centrality–based clutter policy, we again model the environment as a navigation graph over reachable coordinates and compute each node’s load centrality, i.e., the amount of shortest-path ``traffic'' routed through that node when considering shortest paths between many pairs of nodes in the graph. 
Intuitively, nodes with high load are those that would carry high expected traversal flow under shortest-path routing. 
We then instantiate clutter by occluding a fixed percentage of nodes sampled according to their load-centrality scores, thereby preferentially blocking high-traffic regions. 
This differs from our standard policy in the main paper, which is betweenness-centrality–based: while betweenness counts how often a node lies on shortest paths (emphasizing structural connectors/bridges), load centrality emphasizes the aggregate amount of routed flow through nodes under shortest-path traffic assumptions, often producing a slightly different emphasis on globally ``busy'' regions even when multiple alternative shortest paths exist. \\

\noindent The results shown in \cref{tab:abl_cl_gen} follow the same qualitative conclusion as the betweenness-centrality setting in the main paper: CoReLIN (known) consistently outperforms the strongest baselines across floorplan sizes under load-centrality–based clutter. 
CoReLIN achieves near-ceiling success rates while maintaining higher efficiency, particularly as environments scale (e.g., stronger LES in $4-6$ rooms and competitive best performance in $7-10$ rooms), whereas Always Interact / Always Detour / Clean + S/P degrade more substantially in larger layouts. 
Overall, these findings reinforce that our improvements are robust to the choice of clutter generation policy, and not specific to the betweenness-based clutter distribution used in the main paper.

\subsection{Between-ness centrality based}
We next evaluate our primary clutter generation policy -- betweenness-centrality based -- where clutter is preferentially placed on graph nodes that lie on many shortest paths, thereby occluding structurally important connectors and inducing bottlenecks. 
Our approach (CoReLIN) achieves the strongest overall performance across the reported metrics, while prior interactive-navigation baselines (e.g., ADIN, InterNav, FastDownward) degrade substantially under this more adversarial, connectivity-targeted clutter distribution. 
As expected, simple rule-based strategies such as Always Interact and Always Detour exhibit competitive behavior in some regimes but remain consistently below our method, reinforcing that the gains persist even when clutter is concentrated on globally critical parts of the environment graph.

\section{Failure Analysis}
\label{sec:failure_analysis}

\subsection{LLM Output Formatting Errors}
A recurring failure mode in our approach arises from the LLM occasionally producing responses that do not conform to the predefined output schema required by the downstream planner. 
In such cases, the response cannot be parsed reliably, necessitating an additional query to obtain a valid action specification. 
Although these errors do not typically alter the semantic intent of the predicted action, they introduce avoidable latency and extra querying overhead, which can accumulate over long-horizon episodes. 
We therefore analyze this failure mode separately to quantify its frequency and assess its practical impact on overall system efficiency. \\

\noindent We quantify formatting failures by measuring the fraction of LLM queries that do not conform to the required output schema, along with the resulting re-query and efficiency overhead. Across all evaluated floorplans, the mean invalid-format rate is $\mathbf{0.0805\%}$, indicating that such errors are rare. 
Consequently, the average number of re-queries triggered by formatting failures is only $\mathbf{0.043}$ per episode, corresponding to roughly one additional retry every $23$ episodes. 
These failures also impose minimal computational overhead, with an average of $\mathbf{164.8}$ extra tokens and $\mathbf{0.49}$ seconds of additional latency per episode. 
Overall, these results suggest that while formatting errors do occasionally occur, they are infrequent and have negligible impact on the efficiency of long-horizon task execution.

\subsection{Missed Object and Receptacle Discovery}
In this subsection, we analyze the dominant failure mode of our approach in the unknown setting. 
Since low-level manipulation is assumed reliable, episodes do not terminate due to pickup or placement failures; instead, failures arise exclusively when the agent is unable to discover the task-relevant object or the target receptacle within the episode horizon. 
We conduct this analysis to better understand whether failures are driven primarily by missing the manipulable object, missing the receptacle, or by specific physical properties of the missed instances. 
To this end, we report three metrics: (i)~the percentage of failed episodes attributable to not finding the task object, (ii)~the percentage of failed episodes attributable to not finding the receptacle object, and (iii)~a breakdown of failures by object size and category, where size is determined from the simulator-provided bounding-box volume. 
Together, these statistics help identify the types of targets that are most challenging to detect and clarify the perceptual limitations underlying unsuccessful episodes. \\

\noindent We analyze unsuccessful episodes by categorizing the final failure according to whether the agent failed to discover the task-relevant object or the target receptacle. 
Our results indicate that missed task-object discovery is slightly more common, accounting for $48\%$ of failures, compared to $40\%$ for missed receptacle discovery. 
A breakdown by object size further shows a strong asymmetry between the two roles. 
Missed task objects are concentrated entirely among small and medium-sized instances, indicating that portable objects are particularly vulnerable to occlusion and limited observability. 
In contrast, missed receptacles are predominantly large, suggesting that receptacle failures are not driven by scale, but instead by incomplete search coverage or difficulty localizing the correct semantic target within cluttered scenes. 
At the category level, failures most frequently involve task objects such as TissueBox, Mug, RemoteControl, SoapBottle, and Statue, and receptacles such as Desk, GarbageCan, and ShelvingUnit. 
Overall, these results indicate that the dominant failure mode in the unknown setting is imperfect target discovery, with small manipulable objects being the most challenging cases.

\subsection{Manipulation Failures in Real-Robot Experiments}
In real-robot experiments, beyond the perception and exploration failures discussed above, we also observed occasional failures during low-level manipulation execution, particularly during grasping and placement. 
These failures arise after the high-level policy has already selected a correct target object or receptacle, and are therefore attributable to the lower-level manipulation controller rather than to the semantic reasoning module. 
In our system, such failures are handled by re-invoking the manipulation primitive with a newly sampled nearby end-effector pose until success or episode termination. 
We did not systematically annotate these events across all runs, we instead treat them as a qualitative limitation of the deployed robotic stack rather than a quantified failure mode.

\section{Example run-through}
\label{sec:run_through}

\cref{F:rt_1}, \cref{F:rt_2}, and \cref{F:rt_3} illustrate representative decision-making snapshots from CoReLIN across different stages of an episode, showing how the same underlying policy adapts its behavior as the remaining horizon changes. 
We use these examples to concretely demonstrate how the model reasons over the current scene graph, discovered objects, and path structure to choose actions that are appropriate for the episode stage. 
Early in the episode (\cref{F:rt_1}), CoReLIN prioritizes exploration of relevant regions to discover task-relevant objects and receptacles that are still unknown. 
Midway through the episode (\cref{F:rt_2}), once sufficient scene knowledge has been accumulated, it identifies high-impact obstacles and proactively rearranges them when doing so improves future navigation and interaction efficiency. 
Late in the episode (\cref{F:rt_3}), when only a few tasks remain and the opportunity for long-term gains from further rearrangement is limited, CoReLIN instead favors direct task completion by detouring around obstacles rather than spending time optimizing the environment. 
Together, these examples highlight the central principle of our approach: balancing exploration, environmental optimization, and detour-based execution according to the episode’s remaining long-horizon value.

\begin{figure*}[t]
\centering
\includegraphics[width=\linewidth]{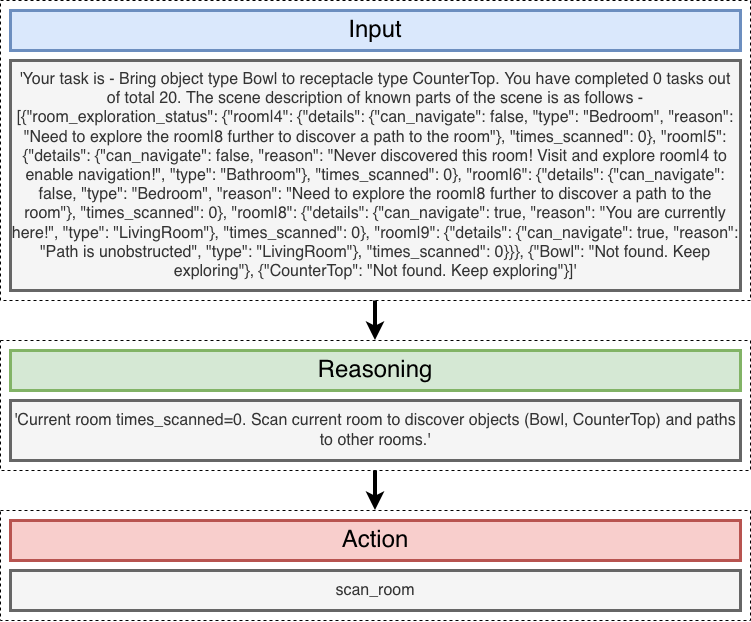}
\caption{\textbf{Early-episode behavior:} At the beginning of an episode, CoReLIN prioritizes exploration to reduce scene uncertainty and discover task-relevant objects, receptacles, and navigation constraints. In this example, because the bowl and countertop have not yet been observed and the current room has not been scanned, the agent reasons that scanning the room is the most valuable action for revealing relevant objects and identifying paths to other rooms, thereby improving future task planning.}\small
\label{F:rt_1}
\end{figure*}

\begin{figure*}[t]
\centering
\includegraphics[width=\linewidth]{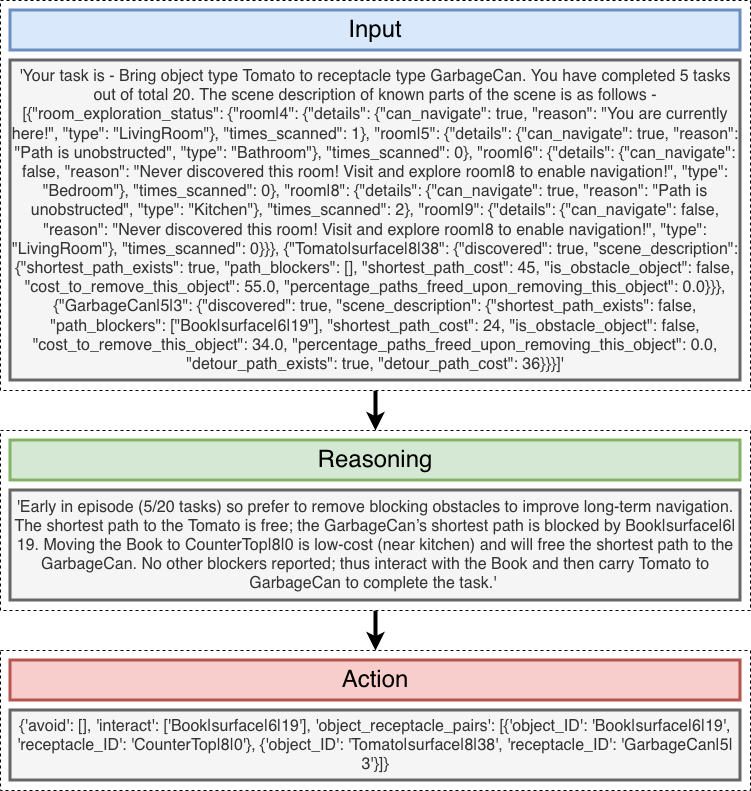}
\caption{\textbf{Mid-episode behavior:} Once sufficient scene structure has been uncovered, CoReLIN shifts from pure exploration to environment optimization, removing high-impact obstacles when doing so yields long-term gains across the remaining tasks. In this example, although the tomato is already reachable, the shortest path to the garbage can is blocked by a book. Since the episode is still relatively early and clearing this obstacle opens the direct route at low cost, the agent first relocates the book and then completes the transport, improving navigability for future task execution.}\small
\label{F:rt_2}
\end{figure*}

\begin{figure*}[t]
\centering
\includegraphics[width=\linewidth]{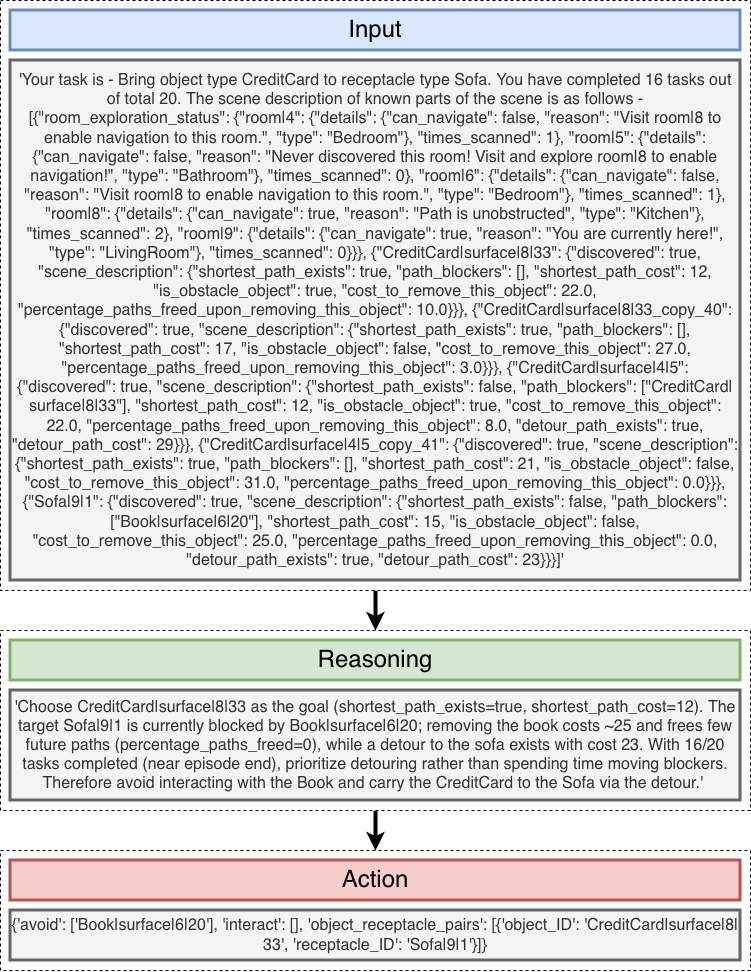}
\caption{\textbf{Late-episode behavior:} Near the end of an episode, CoReLIN prioritizes immediate task completion over further environment optimization, since there is little remaining opportunity to benefit from costly rearrangement. In this example, although the sofa’s shortest path is blocked by a book, removing that obstacle would provide negligible future utility because most tasks have already been completed. Instead, the agent chooses to avoid the blocker and transport the credit card to the sofa via an available detour, reflecting a late-stage preference for efficient completion rather than long-term scene improvement.}\small
\label{F:rt_3}
\end{figure*}

\section{Hardware Deployment}
\label{sec:Spot}
We deploy our method on a Boston Dynamics Spot robot equipped with the Spot Arm, demonstrating that our system transfers from simulation to a real mobile manipulator. Below, we describe the full hardware pipeline used in our experiments.

\subsection{Perception and Object Recognition}
Spot is equipped with five front-facing fisheye cameras and a wrist-mounted depth camera on the end-effector. To remain faithful to our simulation assumptions, we restrict active perception during navigation to only the front fisheye RGB-D streams; the wrist camera is used exclusively during grasping and placement to ensure reliable manipulation.
Object detection on fisheye images is performed in a two-stage process:

\begin{enumerate}
    \item \textbf{Label prediction:} We run a YOLO-based detector to obtain coarse object categories that match our simulation-level abstractions (e.g., \texttt{Mug}, \texttt{Box}, \texttt{Shelf}, \texttt{StudyTable}).

    \item \textbf{Mask generation:} For each detected object, we apply SAM to obtain a high-quality bounding region and mask, which is then projected into 3D using the synchronized depth image.
\end{enumerate}

\noindent Detected objects are inserted into the scene graph, maintaining the same representation used in simulation.

\subsection{Environment Representation and State Estimation}
Since the physical lab layout is fixed and known, we precompute a ground-truth grid graph of traversable free space prior to the episode. 
This graph uses a grid resolution of $1.0$ m, and is used for shortest-path planning, betweenness estimation, and PoC computation. \\

\noindent Spot’s pose within this map is obtained using its onboard state estimator, which fuses IMU, kinematic odometry, and visual-inertial sensing. 
This provides a drift-minimized, metrically consistent localization signal throughout the episode, enabling us to register detections, depth maps, and manipulated-object trajectories directly into the global coordinate frame. \\

\noindent During execution, the robot maintains an estimate of currently reachable nodes. 
For any grid node, we evaluate the local 3D occupancy by analyzing the point cloud density inside a spherical region of fixed radius. 
Let $S(v)$ denote the set of depth points within the sphere centered at node $v$. 
A node is marked \emph{occupied} if $\lvert S(v) \rvert > \tau$ and \emph{free} otherwise. 
We set the occupancy threshold to $\tau = 400$, selected empirically to balance noise suppression and spatial sensitivity. \\

\noindent This online occupancy classification yields a dynamically evolving traversability graph that the LLM planner queries when reasoning about shortest paths, obstacle-induced bottlenecks, and candidate regions for exploration.

\subsection{Low-Level Control and Execution}
All locomotion and manipulation commands are executed using the official \verb|bosdyn| Python SDK. Our system interfaces with Spot via:

\begin{itemize}
    \item \textbf{Navigation:} issuing waypoint-based locomotion goals generated by the mapping module,

    \item \textbf{Manipulation:} using the SDK's grasp and arm-trajectory APIs for pickup and placement.
\end{itemize}

The low-level controller provides robust execution primitives such as:

\begin{itemize}
    \item grasp acquisition with automatic force adjustment,

    \item hand-equipped alignment during approach,

    \item arm release and retreat after placement.
\end{itemize}

\noindent These behaviors abstract away most failures at the control layer, allowing the high-level LLM planner to focus on reasoning about obstacle configuration, exploration, and task sequencing.

\subsection{Runtime Integration}
At each control cycle, the system:

\begin{enumerate}
    \item captures front-camera RGB-D observations,
    
    \item runs YOLO+SAM to update scene graph entries,

    \item updates the reachable grid nodes using depth-based occupancy,

    \item queries the LLM for the next high-level action (navigate, pick-and-place, or explore),

    \item executes the corresponding low-level command through the \verb|bosdyn| API.
\end{enumerate}

\noindent This forms a tight perception–reasoning–action loop that closely mirrors our simulator setup while operating under real sensor noise, partial observability, and actuation uncertainty.

\subsection{Real Robot Experiments}
We provide visualization of an episode executed on Spot robot. \cref{F:spot_figs} shows the point cloud and observed free navigable space as seen by the robot.

\begin{figure*}[t]
\centering

\begin{subfigure}[t]{0.48\linewidth}
    \centering
    \includegraphics[width=\linewidth]{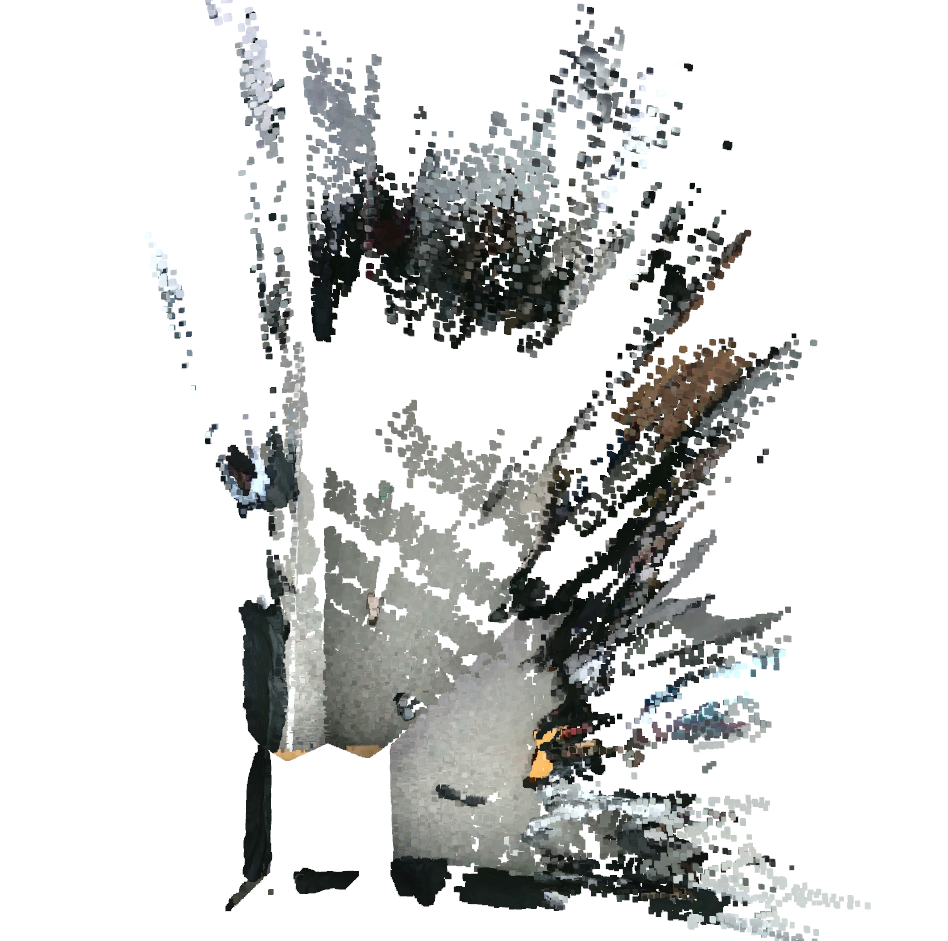}
    \caption{Initial point cloud}
    \label{fig:a}
\end{subfigure}
\hfill
\begin{subfigure}[t]{0.48\linewidth}
    \centering
    \includegraphics[width=\linewidth]{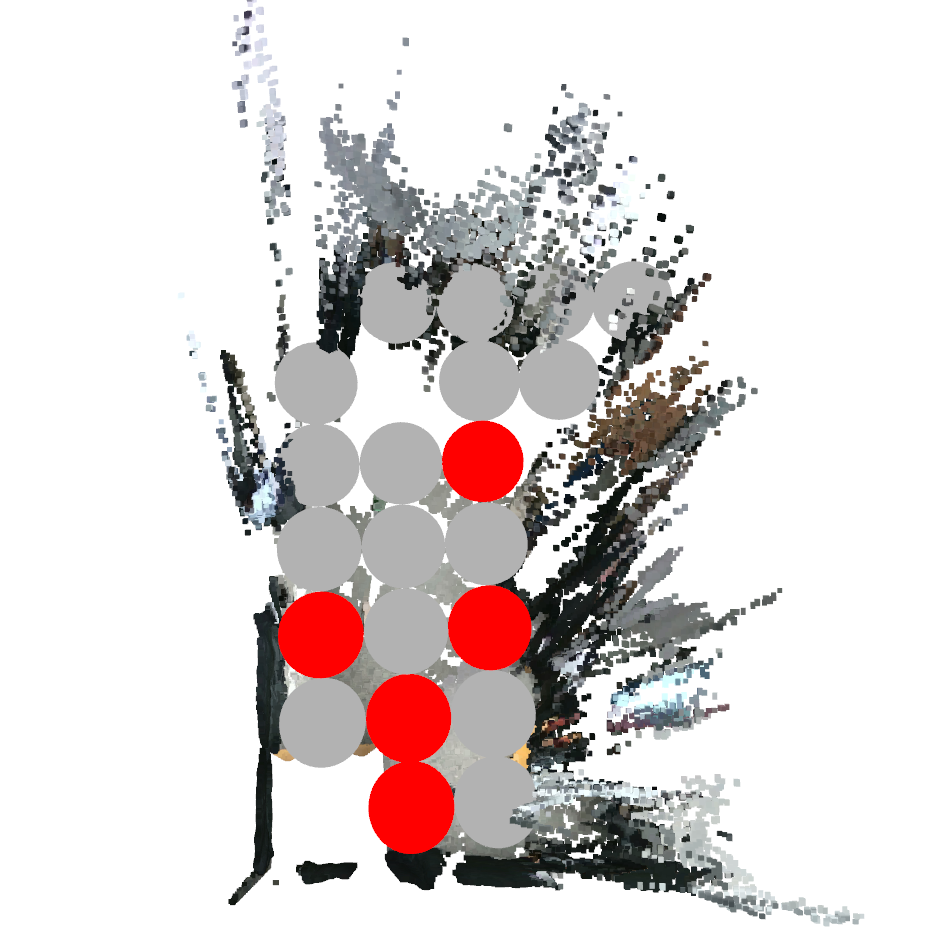}
    \caption{Initial grid graph}
    \label{fig:b}
\end{subfigure}

\medskip

\begin{subfigure}[t]{0.48\linewidth}
    \centering
    \includegraphics[width=\linewidth]{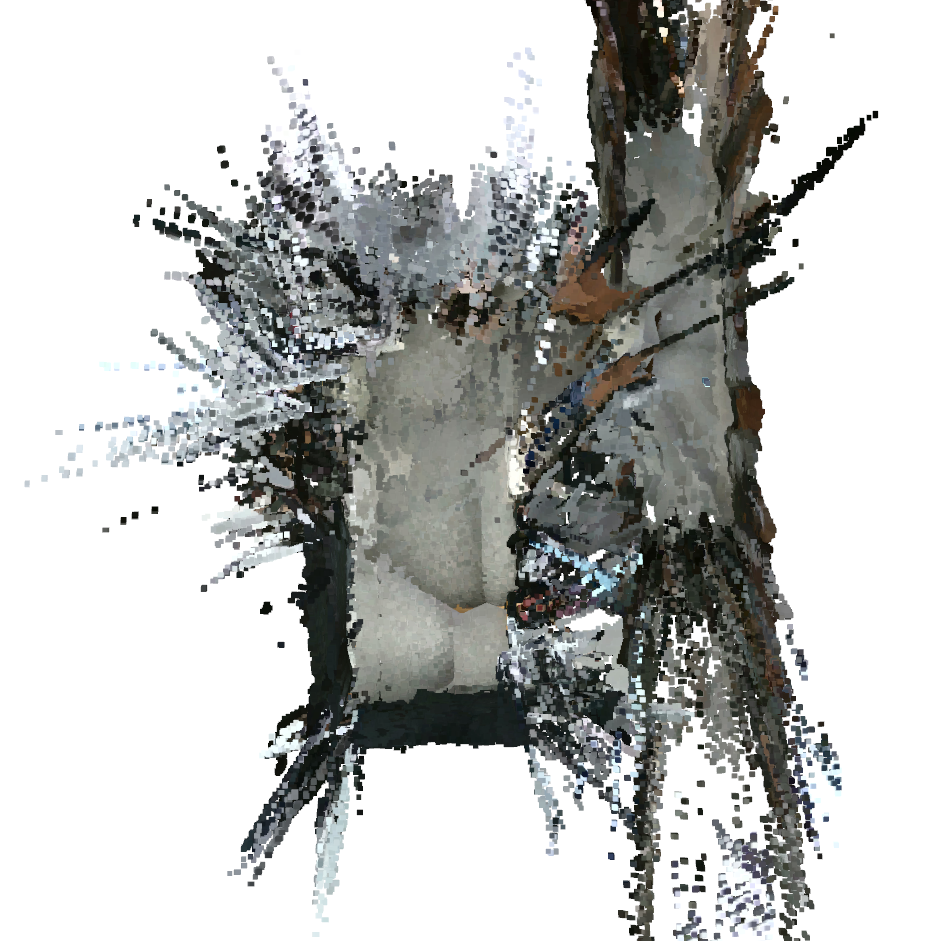}
    \caption{Final point cloud}
    \label{fig:c}
\end{subfigure}
\hfill
\begin{subfigure}[t]{0.48\linewidth}
    \centering
    \includegraphics[width=\linewidth]{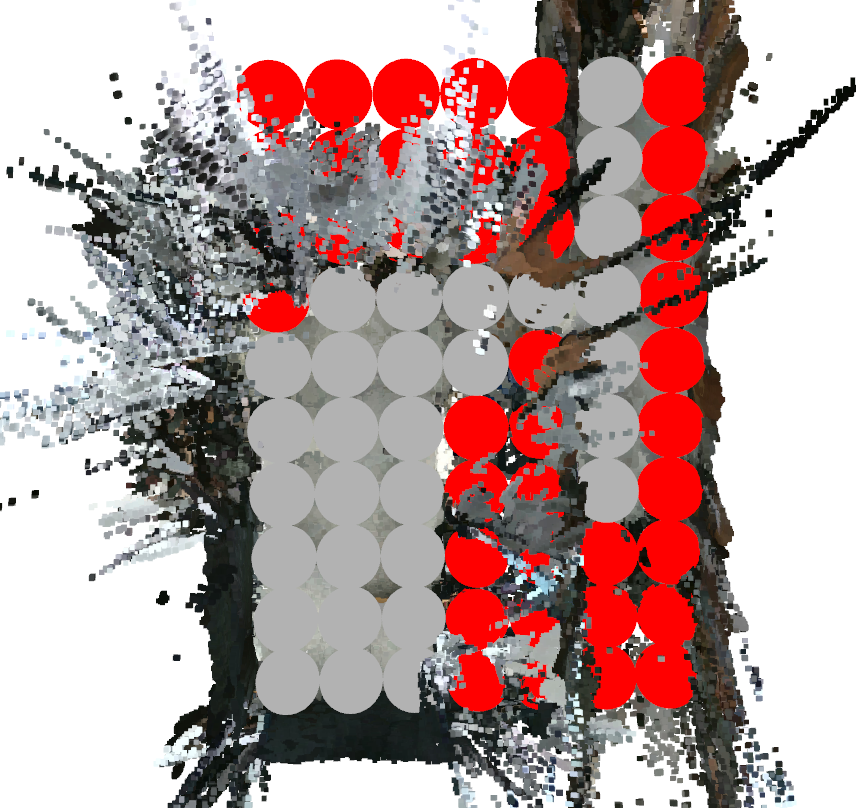}
    \caption{Final grid graph}
    \label{fig:d}
\end{subfigure}

\caption{Point-cloud–driven environment mapping during real-world deployment. We visualize the evolution of the robot’s perception and its derived navigation graph during a real episode on the Boston Dynamics Spot. (a) The initial RGB-D observations from the front fisheye cameras produce a sparse and noisy point cloud reflecting partial scene coverage. (b) From this point cloud, the system constructs an initial navigation grid graph by projecting points onto the 2D floor plane and classifying nodes as free (gray) or occupied (red) based on local point density and presence of obstacles. (c) As the robot actively explores and executes navigation and manipulation actions, the accumulated point cloud becomes significantly more complete, revealing previously unseen regions and freed regions due to obstacle relocation. (d) The final grid graph incorporates these expanded observations, yielding a more accurate and globally connected representation of traversable space and obstacles. This perception-mapping loop provides the structured scene graph used by our LLM planner to reason about exploration, obstacle relocation, and long-horizon navigation decisions.}
\label{F:spot_figs}
\end{figure*}

\section{Prompts}
\label{sec:LLM_prompts}

To facilitate full transparency and reproducibility of our method, we provide the exact system-level (\cref{F:prompt_sys}) and developer-level (\cref{F:prompt_dev}) prompts used to guide the LLM planner in our framework for constrained-based planning. 
Our prompts precisely encode the rules of our setting -- long-horizon, sequential-tasks, partially observed interactive navigation -- and expose to the model only those abstractions available to the robot at test time (e.g., scene-graph structure, obstacle costs, path characteristics, and remaining task budget). 
By doing so, the prompts serve as an explicit interface between high-level reasoning and low-level execution: they constrain the LLM to operate within the capabilities of the underlying planner while enabling flexible, global decision-making about exploration, obstacle relocation, and long-term environment shaping.

\begin{figure*}[t]
\centering
\includegraphics[width=\linewidth]{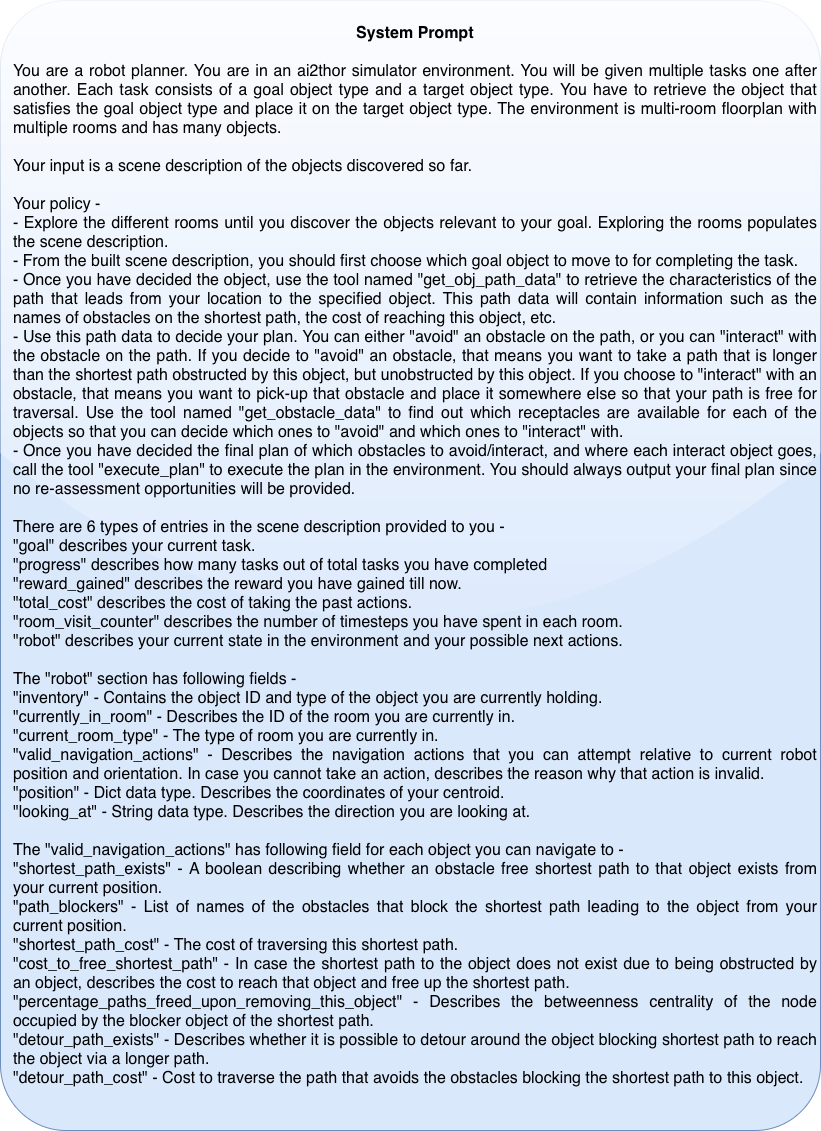}
\caption{Our system prompt for the Large Language Model in our constraint-based planning framework.}\small
\label{F:prompt_sys}
\end{figure*}

\begin{figure*}[t]
\centering
\includegraphics[width=\linewidth]{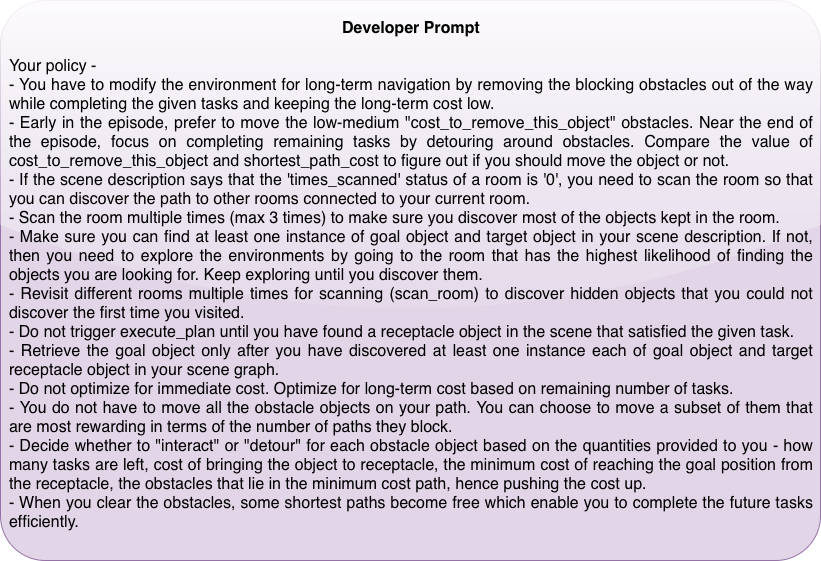}
\caption{Our developer prompt for the Large Language Model in our constraint-based planning framework.}\small
\label{F:prompt_dev}
\end{figure*}

\begin{figure*}[t]
\centering

\begin{subfigure}[t]{\linewidth}
    \centering
    \includegraphics[width=\linewidth]{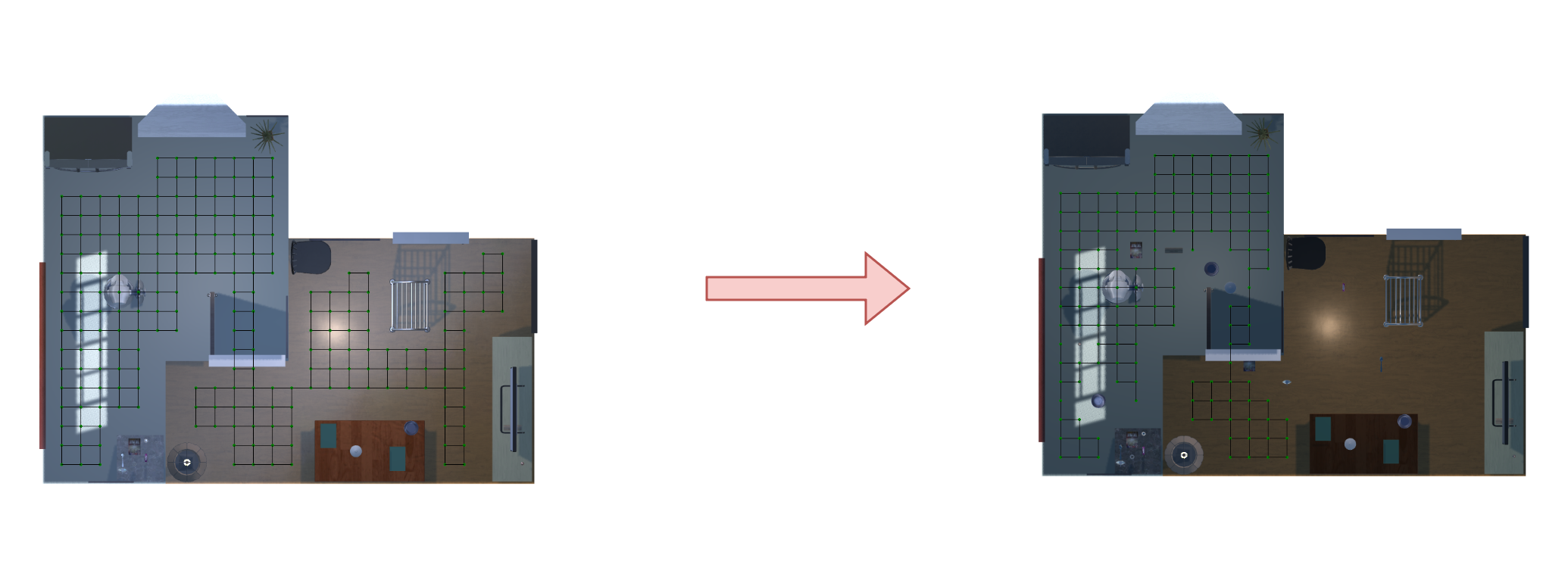}
    \caption{Occlusion in room $1$ floorplan}
    \label{fig:room_1}
\end{subfigure}

\medskip

\begin{subfigure}[t]{\linewidth}
    \centering
    \includegraphics[width=\linewidth]{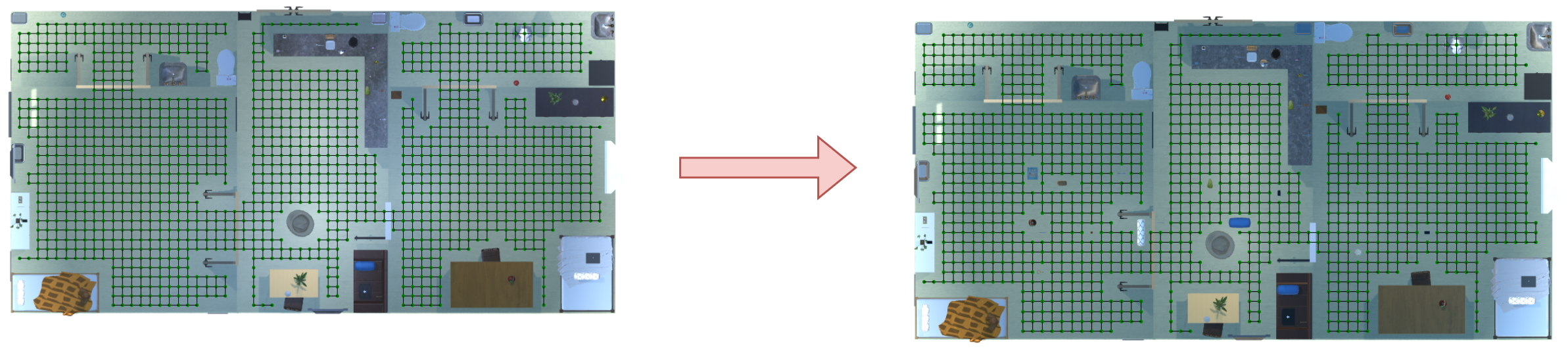}
    \caption{Occlusion in room $5$ floorplan}
    \label{fig:room_5}
\end{subfigure}

\medskip

\begin{subfigure}[t]{\linewidth}
    \centering
    \includegraphics[width=\linewidth]{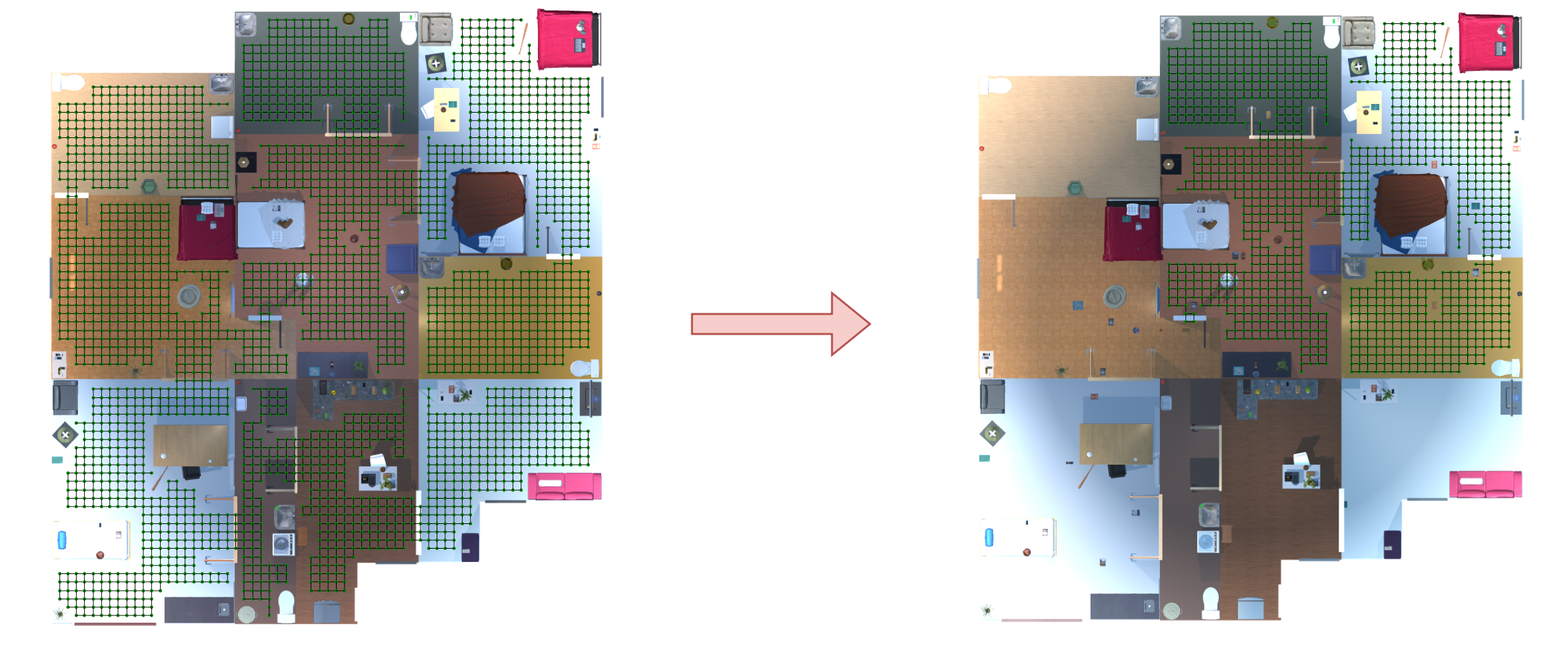}
    \caption{Occlusion in room $10$ floorplan}
    \label{fig:room_10}
\end{subfigure}
\caption{}
\label{F:app_data_viz}
\end{figure*}

\end{document}